\documentclass[10pt]{article} 
\usepackage[preprint]{tmlr}
 
\usepackage{amsmath,amsfonts,bm}

\def\eqref#1{equation~\ref{#1}}

\def\1{\bm{1}}

\DeclareMathAlphabet{\mathsfit}{\encodingdefault}{\sfdefault}{m}{sl}
\SetMathAlphabet{\mathsfit}{bold}{\encodingdefault}{\sfdefault}{bx}{n}

\usepackage{hyperref}
\usepackage{url}
\usepackage{comment}
\usepackage{enumitem}
\usepackage{graphicx}
\usepackage{array}
\usepackage{colortbl}
\usepackage{booktabs}
\usepackage{multirow}
\usepackage{calc}

\usepackage[edges]{forest}
\usepackage{tikz}
\usetikzlibrary{trees, positioning, shapes.geometric, arrows.meta}
\newcolumntype{e}[1]{>{\centering\arraybackslash}p{#1}}

\usepackage{wrapfig}
\usepackage{diagbox}

\definecolor{darkblue}{rgb}{0, 0, 0.5}
\definecolor{hidden-draw}{RGB}{177, 177, 177}
\hypersetup{
    colorlinks=true,
    linkcolor=darkblue,
    citecolor=darkblue,
    pdfpagemode=FullScreen,
}

\title{A Survey on the Honesty of Large Language Models}

\author{Siheng Li$^{1 \dagger}$\thanks{\ Corresponding to: \texttt{shli@se.cuhk.edu.hk}.} \quad Cheng Yang$^{1 \dagger}$ \quad Taiqiang Wu$^2$\thanks{\ Equal contribution.} \\\\
Chufan Shi$^3$ \quad Yuji Zhang$^4$ \quad Xinyu Zhu$^5$ \quad Zesen Cheng$^6$ \quad Deng Cai \quad Mo Yu$^7$ \\\\
Lemao Liu$^7$ \quad Jie Zhou$^7$ \quad Yujiu Yang$^3$ \quad Ngai Wong$^2$ \quad Xixin Wu$^1$ \quad Wai Lam$^1$ \\\\
$^1$The Chinese University of Hong Kong \quad
$^2$The University of Hong Kong \\\\
$^3$Tsinghua University \quad
$^4$University of Illinois at Urbana-Champaign \\\\
$^5$University of Virginia  \quad 
$^6$Peking University \quad
$^7$WeChat AI 
}

\NewDocumentCommand{\yuji}
{ mO{} }{\textcolor{purple}{\textsuperscript{\textit{Yuji}}\textsf{\textbf{\small[#1]}}}}

\NewDocumentCommand{\siheng}
{ mO{} }{\textcolor{blue}{\textsuperscript{\textit{Siheng}}\textsf{\textbf{\small[#1]}}}}

\newcommand\encircle[2][]{\tikz[overlay]\node[fill=blue!20,inner sep=2pt, anchor=text, rectangle, rounded corners=1.5mm,#1] {#2};\phantom{#2}}
\definecolor{known}{RGB}{173, 209, 145}
\definecolor{unknown}{RGB}{166, 166,166}

\newcommand{\known}{\encircle[known, text=white]{K}}
\newcommand{\unknown}{\encircle[unknown, text=white]{U}}

\def\month{MM}  
\def\year{YYYY} 
\def\openreview{\url{https://openreview.net/forum?id=XXXX}} 

\begin{document}

\maketitle

\begin{abstract}
Honesty is a fundamental principle for aligning large language models (LLMs) with human values, requiring these models to recognize what they know and don't know and be able to faithfully express their knowledge. Despite promising, current LLMs still exhibit significant dishonest behaviors, such as confidently presenting wrong answers or failing to express what they know. In addition, research on the honesty of LLMs also faces challenges, including varying definitions of honesty, difficulties in distinguishing between known and unknown knowledge, and a lack of comprehensive understanding of related research. To address these issues, we provide a survey on the honesty of LLMs, covering its clarification, evaluation approaches, and strategies for improvement. Moreover, we offer insights for future research, aiming to inspire further exploration in this important area.

\end{abstract}
\section{Introduction}
Honesty has become a prominent and frequently discussed topic in the development of large language models (LLMs) \citep{askell2021general, bai2022training, touvron2023llama, zhang2023siren, sun2024trustllm}, and is recognized as one of the key principles for aligning LLMs with human preferences and values \citep{askell2021general}. Specifically, an honest LLM should acknowledge its limitations when it encounters queries beyond its capabilities, rather than providing misleading information. This is particularly important in high-stakes domains such as medicine \citep{thirunavukarasu2023large}, law \citep{dahl2024large}, and finance \citep{li2023large}. Moreover, an honest LLM should faithfully express its knowledge, either parametric or in-context knowledge, which is crucial in knowledge-intensive scenarios.

Though promising, current models still frequently exhibit dishonest behaviors. They almost always speak with a confident tone, even when they make errors; they might ``know'' the answer internally but fail to ``say'' it accordingly \citep{li2024inference}; and they may provide biased information influenced by human input \citep{sharma2024towards}. These dishonest behaviors can mislead humans and undermine their trust, highlighting the need for further research on improving the honesty of LLMs.


\tikzstyle{my-box}=[
	rectangle,
	draw=hidden-draw,
	rounded corners,
	text opacity=1,
	minimum height=1.5em,
	minimum width=5em,
	inner sep=2pt,
	align=center,
	fill opacity=.5,
	line width=0.8pt,
]

\tikzstyle{leaf}=[my-box, minimum height=1.5em,
	fill=hidden-pink!80, text=black, align=left,font=\normalsize,
	inner xsep=2pt,
	inner ysep=4pt,
	line width=0.8pt,
]

\begin{figure*}[t]
	\centering
	\resizebox{0.93\textwidth}{!}{ 
        \begin{forest}
			forked edges,
			for tree={
				grow=east,
				reversed=true,
				anchor=base west,
				parent anchor=east,
				child anchor=west,
				base=center,
				font=\large,
				rectangle,
				draw=hidden-draw,
				rounded corners,
				align=center,
                minimum width=4em,
				edge+={darkgray, line width=1pt},
				s sep=3pt,
				inner xsep=2pt,
				inner ysep=3pt,
				line width=0.8pt,
				ver/.style={rotate=90, child anchor=north, parent anchor=south, anchor=center},
			},
			where level=1{text width=11em,font=\normalsize, text centered}{},
			where level=2{text width=11em,font=\normalsize, text centered}{},
			where level=3{text width=12em,font=\normalsize}{},
			where level=4{text width=31em,font=\normalsize, align=left}{},
			[
                \textbf{A Survey on the Honesty of Large Language Models}, ver
                [
					\textbf{Honesty in LLMs}  (\S \ref{sec:honesty_in_llms}), fill=blue!10
			        [                 
						\textbf{What is Honesty in } \\ \textbf{LLMs}  (\S \ref{sec:what_is_honesty_in_llms}), fill=blue!10
                        [
                        \cite{askell2021general,kadavath2022language, evans2021truthful, lin2022teaching,yang2023alignment} \\
                        \cite{chern2024behonest}, align=left, text width=44.5em
                        ]
                    ]
                    [                 
						\textbf{Self-knowledge} (\S \ref{sec:defintion-self-know}), fill=blue!10
                        [
                        \cite{gertler2010self, askell2021general,kadavath2022language, liu2023cognitive,cheng2024can} \\
                        \cite{chern2024behonest,yang2023alignment,zhang2024r, lin2022teaching,xu2024sayself} \\
                        \cite{stengel2024lacie, feng-etal-2024-dont,wen2024art, yadkori2024mitigating}\\
                        \cite{tomani2024uncertainty,xiao2021hallucination,manakul2023selfcheckgpt, kossen2024semantic} \\
                        \cite{chen2024inside,farquhar2024detecting,jiang2023active,wang2023self, ni2024llms} \\
                        \cite{yao2024seakr, yue2024large, jung2024trust,gupta2024language,ramirez2024optimising}, 
                        align=left, text width=44.5em
                        ]
                    ]       
                    [                 
						\textbf{Self-expression} (\S \ref{sec:defintion-self-express}), fill=blue!10
                        [
                        \cite{zhu2023physics, zhang2024how, li2024inference,liu2024lost,shi2024trusting} \\
                        \cite{bai2024hallucination, gu2022robustness, mizrahi2023state, wang2023robustness,sclar2024quantifying} \\
                        \cite{cao2024worst, wei2023simple, sharma2024towards, huang2024trustllm}, 
                        align=left, text width=44.5em
                        ]
                    ]
                ]
                [
                    \textbf{Evaluation of LLM} \\ \textbf{Honesty}  (\S \ref{sec:evaluation_of_llm_honesty}), fill=yellow!10
                    [
                        \textbf{Self-knowledge}  (\S \ref{sec:eval_self_knowledge}), fill=yellow!10
					  [ 
						\textbf{Recognition of} \\ \textbf{Known/Unknown}, fill=yellow!10, text centered
                            [
                            \cite{yin2023large, amayuelas2023knowledge, liu2024examining} \\
                            \cite{gao2024best, chern2024behonest, rajpurkar2016squad} \\
                            \cite{cheng2024can}
                            ]
    					]
    					[	
    					\textbf{Calibration}, fill=yellow!10, text centered	
                            [
                            \cite{guo2017calibration,tian2023just,geng2024survey} \\
                            \cite{zhu2023calibration,xiong2024can, lyu2024calibrating, brier1950verification} \\
                            \cite{naeini2015obtaining, nixon2019measuring,kull2019beyond} \\
                            \cite{degroot1983comparison}
                            ]
					  ]
                        [	
    					\textbf{Selective Prediction}, fill=yellow!10, text centered	
                            [
                            \cite{ren2023outofdistribution,xiong2024can,chen2023adaptation} \\
                            \cite{tomani2024uncertainty, xu2024sayself, mahaut2024factual} \\
                            \cite{ren2023self,geng2024survey,lin2024generating} \\ 
                            \cite{si2023getting}
                            ]
					  ]
                    ]
                    [
                        \textbf{Self-expression}  (\S \ref{sec:eval_self_expression}), fill=yellow!10
					  [ 
						\textbf{Identification-based} \\ \textbf{Evaluation}, fill=yellow!10, text centered
                            [
                            \cite{yin2023large, amayuelas2023knowledge, liu2024examining} \\
                            \cite{gao2024best, chern2024behonest, rajpurkar2016squad} \\
                            \cite{cheng2024can}
                            ]
    					]
    					[	
    					\textbf{Identification-free} \\ \textbf{Evaluation}, fill=yellow!10, text centered	
                            [
                            \cite{sclar2024quantifying,chern2024behonest,mizrahi2023state} \\
                            \cite{cao2024worst, huang2024trustllm, wei2023simple} \\
                            \cite{li2024benchmarking}
                            ]
					  ]
                    ]   
                ]
                [
                    \textbf{Improvement of } \\ \textbf{Self-knowledge}  (\S \ref{sec:improvement_of_self_knowledge}), fill=orange!10
                    [
                        \textbf{Training-free} \\ \textbf{Approaches}  (\S \ref{sec:Improvement_of_self_knowledge_training_free}), fill=orange!10
					  [ 
						\textbf{Predictive Probability}, fill=orange!10, text centered
                            [
                            \cite{xiao2022uncertainty, kadavath2022language, kuhn2023semantic} \\
                            \cite{adiwardana2020towards, malinin2020uncertainty, si2022prompting} \\
                            \cite{ren2023self, duan2024shifting}
                            ]
    					]
    					[	
    					\textbf{Prompting}, fill=orange!10, text centered	
                            [
                            \cite{kadavath2022language, zhao2024fact, jiang2024self} \\
                            \cite{yin2023large,tian2023just,xiong2024can} \\
                            \cite{zhou2023navigating} 
                            ]
					  ]
                        [	
    					\textbf{Sampling and} \\ \textbf{Aggregation}, fill=orange!10, text centered	
                            [
                            \cite{xiong2024can, yang2024just, zhou2022prompt} \\
                            \cite{lyu2024calibrating, kuhn2023semantic, lin2024generating} \\
                            \cite{manakul2023selfcheckgpt, yadkori2024mitigating, huang2024calibrating} \\
                            \cite{farquhar2024detecting, fadeeva2024fact}
                            ]
					  ]
                    ]
                    [
                        \textbf{Training-based} \\ \textbf{Approaches}  (\S \ref{sec:improvement_of_self_knowledge_training_based}), fill=orange!10
					  [ 
						\textbf{Supervised} \\ \textbf{Fine-tuning}, fill=orange!10, text centered
                            [
                            \cite{yang2023alignment, zhang2024r, cheng2024can} \\
                            \cite{chen2024teaching, wan2024knowledge, kapoor2024large} \\
                            \cite{lin2022teaching, ulmer2024calibrating, han2024enhancing}
                            ]
    					]
    					[	
    					\textbf{Reinforcement} \\ \textbf{Learning}, fill=orange!10, text centered	
                            [
                            \cite{cheng2024can, xu2024rejection, gao2024best} \\
                            \cite{xu2024sayself, stengel2024lacie, band2024linguistic}
                            ]
					  ]
                        [	
    					\textbf{Probing}, fill=orange!10, text centered	
                            [
                            \cite{kadavath2022language, azaria2023internal} \\
                            \cite{marks2023geometry,liu2024universal, burns2023discovering} \\
                            \cite{kossen2024semantic, ji2024llm}
                            ]
					  ]
                    ]   
                ]
                [
                    \textbf{Improvement of } \\ \textbf{Self-expression}  (\S \ref{sec:improvement_of_self_expression}), fill=orange!10
                    [
                        \textbf{Training-free} \\ \textbf{Approaches}  (\S \ref{sec:improvement_of_imporving_self_consistency_training_free}), fill=orange!10
					  [ 
						\textbf{Prompting}, fill=orange!10, text centered
                            [
                            \cite{wei2022chain, kojima2022large, zhou2023leasttomost} \\
                            \cite{press2023measuring, zheng2024take, wang2023plan} \\
                            \cite{zhao2024fact}
                            ]
    					]
    					[	
    					\textbf{Decoding-time} \\ \textbf{Intervention}, fill=orange!10, text centered	
                            [
                            \cite{li2024inference, chen2024context, chuang2024dola} \\
                            \cite{zhang2023alleviating, shi2024trusting, leng2024mitigating}
                            ]
					  ]
                        [	
    					\textbf{Sampling and} \\ \textbf{Aggregation}, fill=orange!10, text centered	
                            [
                            \cite{wang2023selfconsistency, chen2023universal, wang2024integrate} \\
                            \cite{thirukovalluru2024atomic}
                            ]
					  ]
                        [	
    					\textbf{Post-generation} \\ \textbf{Revision}, fill=orange!10, text centered	
                            [
                            \cite{dhuliawala2023chain, varshney2023stitch, zhao2023verify} 
                            ]
					  ]
                    ]
                    [
                        \textbf{Training-based} \\ \textbf{Approaches}  (\S \ref{sec:improvement_of_imporving_self_consistency_training_based}), fill=orange!10
                        [ 
						\textbf{Self-aware} \\ \textbf{Fine-tuning}, fill=orange!10, text centered
                            [
                            \cite{yang2023alignment, zhang2024r, cheng2024can} \\
                            \cite{wan2024knowledge, kang2024unfamiliar}
                            ]
    					]
					  [ 
						\textbf{Self-supervised} \\ \textbf{Fine-tuning}, fill=orange!10, text centered
                            [
                            \cite{tian2024finetuning, zhang2024self, lin2024flame}
                            ]
    					]
                    ]  
                ]
                [
					\textbf{Future Work} (\S \ref{sec:future_work}), fill=green!10
                    [
                        Objective or Subjective{,}
                        Knowledge Identification{,} Honesty in Instruction-following{,} \\ Honesty on In-context Knowledge{,} Honesty in Various Models,
						text width=46em, fill=green!10
                    ]
				]
			]
		\end{forest}
	}
	\caption{The outline of this survey.}
    \label{outline}
\end{figure*}
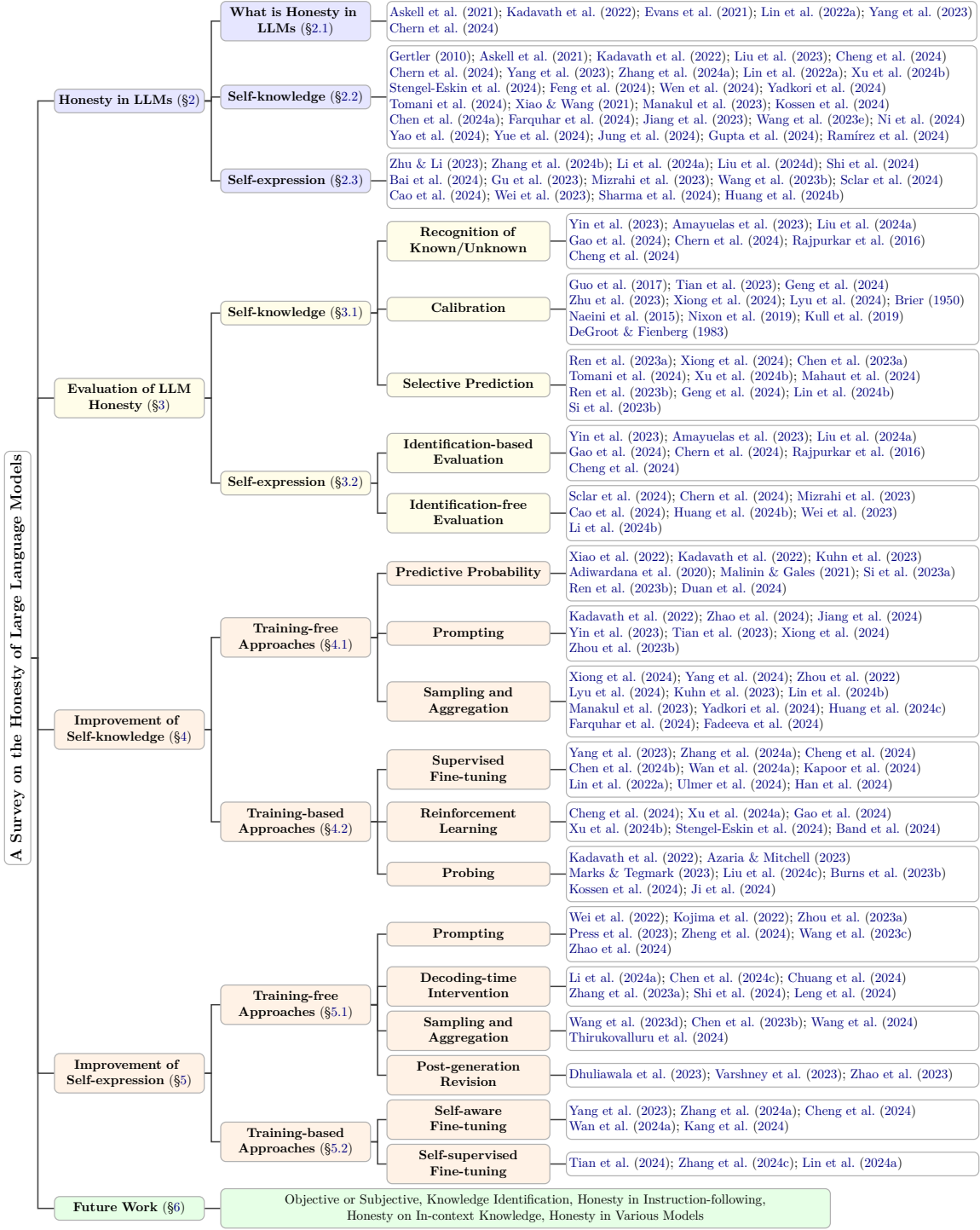

However, research on the honesty of LLMs also faces several challenges. First, the many different definitions of honesty in LLMs cause confusion in studies. Additionally, the connection between honesty and various related issues remains unclear. Second, honesty is specific to each model, as it requires identifying the model’s known and unknown knowledge, making both its evaluation and improvement challenging. Last but not least, although many studies address related aspects of honesty, such as recognizing known and unknown information \citep{yin2023large}, verbalizing confidence \citep{tian2023just}, and fine-tuning based on models’ internal knowledge \citep{zhang2024r}, there is a lack of comprehensive understanding of these studies, which could potentially foster mutual benefits among them.

To address the aforementioned challenges and promote further research on the honesty of LLMs, we provide an extensive overview of current studies in this area. Figure \ref{outline} shows the outline of this survey. We start by summarizing the widely accepted and inclusive definitions on the honesty of LLMs from previous research (\textsection\ref{sec:honesty_in_llms}).
Next, we introduce existing evaluation approaches for assessing the honesty of LLMs (\textsection\ref{sec:evaluation_of_llm_honesty}). 
We then offer an in-depth review of research focused on improving the honesty of LLMs (\textsection\ref{sec:improvement_of_self_knowledge}, \textsection\ref{sec:improvement_of_self_expression}). Finally, we propose potential directions for future research on the honesty of LLMs (\textsection\ref{sec:future_work}). We will constantly update the related research at \url{https://github.com/SihengLi99/LLM-Honesty-Survey}.

\section{Honesty in LLMs}
\label{sec:honesty_in_llms}

\begin{figure}[t]
    \centering
    \includegraphics[width=\linewidth]{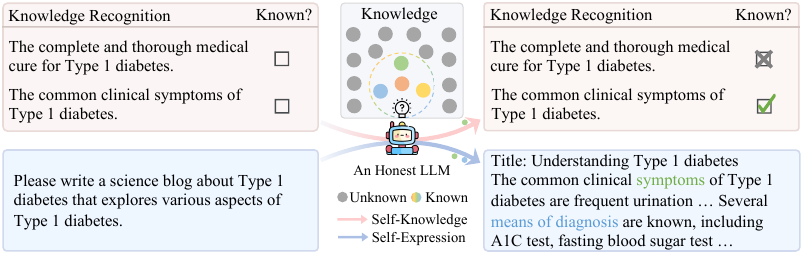}
    \caption{An illustration of an honest LLM that demonstrates both self-knowledge and self-expression.}
    \label{fig:honesty}
\end{figure}

\subsection{What is Honesty in LLMs}
\label{sec:what_is_honesty_in_llms}

Conceptually, being honest is described as being ``free of deception, morally upright or virtuous, among other traits'' \citep{dictionary1989oxford}. In the realm of LLMs, there has been a long-standing pursuit of developing honest models, with various definitions of honesty emerging over time. \citet{askell2021general} describe honesty as providing accurate information, expressing uncertainty without misleading, and being aware of knowledge and internal state. \citet{kadavath2022language} indicate honesty as an umbrella concept including truthfulness, calibration, self-knowledge, explainability, and non-deceptiveness. In simpler terms, researchers consider a model as honest if it refrains from making statements it doesn't believe \citep{evans2021truthful} or if it is able to express everything represented in its internal states through natural language \citep{lin2022teaching}. Recent studies suggest that an honest model should accurately express its knowledge and humbly acknowledge its limitations without deception or being inconsistent \citep{yang2023alignment, chern2024behonest}.

To summarize, the most widely accepted definitions for an honest LLM are \textit{self-knowledge} and \textit{self-expression}. Self-knowledge involves the model being aware of its own capabilities, recognizing what it knows and what it doesn't, allowing it to acknowledge limitations or convey uncertainty when necessary. Self-expression refers to the model's ability to faithfully express its knowledge, leading to reliable outputs. In this paper, we consider an LLM to be honest if it fulfills these two widely accepted criteria: \textit{possessing both self-knowledge and self-expression}. An illustrated example is shown in Fig. \ref{fig:honesty}, with detailed explanations provided below. 

\subsection{Self-knowledge}
\label{sec:defintion-self-know}
The concept of self-knowledge is crucial in both the philosophy of mind and epistemology, referring to one's understanding of their own mental states, such as experience, thoughts, beliefs, and desires \citep{gertler2010self}. Within the context of LLMs, research on self-knowledge has also emerged as a prominent and rapidly growing field of interest \citep{askell2021general, kadavath2022language, liu2023cognitive, cheng2024can, chern2024behonest}. Specifically, the self-knowledge capacity of LLMs hinges on their ability to recognize what they know and what they don’t know. This enables them to explicitly state \textit{``I don't know''} when lacking necessary knowledge, thereby avoiding making wrong statements \citep{yang2023alignment, zhang2024r}. Additionally, it also allows them to provide confidence or uncertainty\footnote{In this paper, We do not explicitly distinguish between confidence and uncertainty, as they are two sides of the same coin: when confidence increases, uncertainty decreases \citep{geng2024survey}.} indicators in responses to reflect the likelihood of their correctness \citep{lin2022teaching, xu2024sayself, stengel2024lacie}.

Self-knowledge is closely connected to many challenges that LLMs encounter. For example, empowering LLMs with the ability to refuse answering unknown questions can help mitigate hallucinations \citep{feng-etal-2024-dont, wen2024art, yadkori2024mitigating, tomani2024uncertainty}. In addition, an LLM's uncertainty can serve as a valuable indicator for detecting hallucinations \citep{xiao2021hallucination, manakul2023selfcheckgpt, kossen2024semantic, chen2024inside, farquhar2024detecting}. Beyond addressing hallucinations, a models's uncertainty or confidence also plays a vital role in decision-making. For instance, it can determine when external knowledge is needed in adaptive retrieval augmentation \citep{jiang2023active, wang2023self, liu2024ctrla, ni2024llms, yao2024seakr}, or whether it is necessary to invoke another LLM in a model cascading scenario \citep{yue2024large, jung2024trust, gupta2024language, ramirez2024optimising}.

\subsection{Self-expression}
\label{sec:defintion-self-express}
In human society, self-expression involves conveying one’s thoughts and feelings through languages, decisions, or actions, and it is regarded as a highly respected value in Western civilization \citep{kim2011culture}. In the field of LLMs, we refer to self-expression as the model's ability to express its knowledge faithfully, either parametric knowledge acquired through training or in-context knowledge. This enables the model to ground its responses in its knowledge rather than fabricating information.
 
Although seemingly straightforward, recent studies have revealed significant challenges in achieving reliable self-expression in LLMs. For example, \citet{zhu2023physics, zhang2024how, li2024inference} highlight that even when LLMs possess the knowledge internally, they may not be able to express it accurately. Additionally, these models often struggle to accurately convey in-context knowledge, such as relevant articles or figures \citep{liu2024lost, shi2024trusting, bai2024hallucination, zhang2024knowledge}. Both these issues can contribute to the occurrence of hallucinations. Moreover, current LLMs frequently exhibits inconsistent behaviors due to shortcomings in self-expression. For instance, slight changes in prompts may lead to substantial performance degradation \citep{gu2022robustness, mizrahi2023state, wang2023robustness, sclar2024quantifying, cao2024worst}, or the model might provide biased information to cater to the user's views \citep{wei2023simple, wang2023can, sharma2024towards, huang2024trustllm}. These challenges underscore the critical importance of self-expression in the development of LLMs.

\section{Evaluation of LLM Honesty}
\label{sec:evaluation_of_llm_honesty}
In this section, we review previous research on the evaluation of honesty and consolidate these efforts into two categories: evaluations of self-knowledge (\textsection\ref{sec:eval_self_knowledge}) and self-expression (\textsection\ref{sec:eval_self_expression}).

\subsection{Self-knowledge}
\label{sec:eval_self_knowledge}
\begin{figure}
    \centering
    \includegraphics[width=\linewidth]{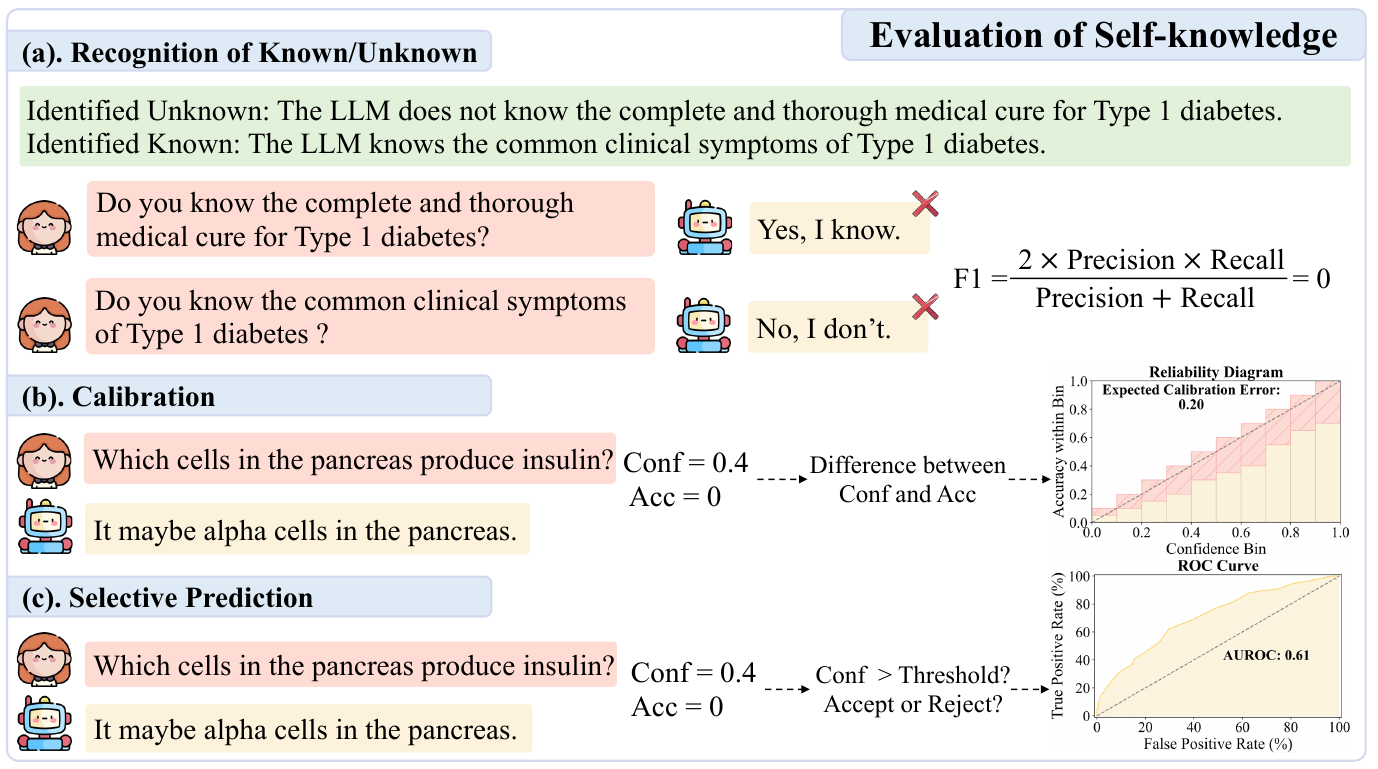}
    \caption{Illustrations of self-knowledge evaluation, encompassing the recognition of known/unknown, calibration, and selective prediction. ``Conf'' indicates the LLM's confidence score and ``Acc'' represents the accuracy of the response.}
    \label{fig:evaluation_self_knowledge}
\end{figure}

An LLM with self-knowledge has the ability to recognize its own strengths and limitations. There are generally two approaches for evaluating self-knowledge. The first is a binary judgement regarding the capacity of LLMs on \textit{recognition of known/unknown}. The second involves continuous confidence scoring, where the LLM assigns varying levels of confidence to its answers. This evaluation includes \textit{calibration} and \textit{selective prediction}. Fig. \ref{fig:evaluation_self_knowledge} provides examples of these assessments.

\paragraph{Recognition of Known/Unknown.}

\begin{table}[t]
\centering
\caption{Model-agnostic benchmarks for recognition of known/unknown. ``\%U'' denotes the proportion of unknown questions. \known\ :Known questions,  \unknown\ :Unknown questions.}
\label{tab:known_unknown_benchmark}
\vspace{0.5\baselineskip}
\resizebox{\textwidth}{!}{
\begin{tabular}{lll}
\toprule
\textbf{Benchmark} & \textbf{Size~(U\%)} & \textbf{Description} \\ \midrule

SelfAware~\citep{yin2023large} & 3369~(31\%) & \begin{tabular}[c]{@{}p{14.5cm}@{}}\known\ are from SQuAD~\citep{rajpurkar2016squad}, HotpotQA~\citep{yang2018hotpotqa} and TriviaQA~\citep{joshi2017triviaqa}; \unknown\ are collected from platforms like Quora and HowStuffWorks and then filtered by humans. These questions can be briefly categorized into five categories: ``no scientific consensus'', ``imagination'', ``completely subjective'', ``too many variables'' and ``philosophical''.\end{tabular} \\ \midrule

KUQ~\citep{amayuelas2023knowledge} & 6884~(50\%) & \begin{tabular}[c]{@{}p{14.5cm}@{}}\known\ are from SQuAD, HotpotQA and TriviaQA; \unknown\ are annotated by crowd-sourced workers according to six categories: ``future unknown'', ``unsolved problem'', ``controversial'', ``w/ false assumption'', ``counterfactual'' and ``ambiguous''.\end{tabular} \\ \midrule

UnknownBench~\citep{liu2024examining} & 13319~(50\%) & \begin{tabular}[c]{@{}p{14.5cm}@{}}\known\ are from TPQ of FalseQA~\citep{hu2023won}, NaturalQuestion~\citep{kwiatkowski2019natural} and template-generated data; \unknown\ are from FPQ of FalseQA, non-existent-concept induced NaturalQuestion and non-existent-concept induced template-generated data.\end{tabular} \\ \midrule

HoneSet~\citep{gao2024best} & 930~(100\%) & \begin{tabular}[c]{@{}p{14.5cm}@{}}\unknown\ are generated by GPT-4 according to five categories and then filtered by human annotators. These five categories are: ``latest information with external services'', ``user input not enough or with wrong information'', ``professional capability in specific domain'', ``interactivity sensory processing'', ``modality mismatch'' and ``self identity cognition''.\end{tabular} \\ \midrule

BeHonest~\citep{chern2024behonest} & 12227~(63\%) & \known\ and \unknown\ are collected from SelfAware and UnknownBench. \\ 

\bottomrule
\end{tabular}%
}
\end{table}
LLMs should be capable of discerning what they know and what they don't, in order to avoid misleading users when they lack relevant information. Current evaluation of this ability can be broadly categorized into two types: model-agnostic \citep{yin2023large, amayuelas2023knowledge, liu2024examining, gao2024best, chern2024behonest} and model-specific \citep{cheng2024can}, depending on whether the approach is tailored to a particular LLM. 

\textit{Model-agnostic} approach applies the same set of known and unknown questions across all LLMs. Representative benchmarks in this category include SelfAware \citep{yin2023large}, KUQ \citep{amayuelas2023knowledge}, UnknownBench \citep{liu2024examining}, HoneSet \citep{gao2024best} and BeHonest \citep{chern2024behonest}. These benchmarks generally assume that the model's pre-training corpus forms its knowledge base. For example, \cite{yin2023large} consider Wikipedia as part of the model's known knowledge as it is othen included in pre-training data. Consequently, questions sourced from Wikipedia (e.g., SQuAD \citep{rajpurkar2016squad}) can be treated as known questions. For unknown questions, a heuristic annotation process is often used. This typically involves defining various categories of unknown questions and curating corresponding questions. For instance, HoneSet \citep{gao2024best} identifies five categories (e.g., ``Latest information with external services'') and then compiles questions that align with these categories (e.g., ``Show the current most-watched movies on Netflix''). Further details on each model-agnostic benchmark can be found in Tab. \ref{tab:known_unknown_benchmark}.

\textit{Model-specific} approach tailors question sets for each LLM. A notable benchmark for this is Idk \citep{cheng2024can}, which distinguishes between known and unknown questions based on the model's performance. Specifically, it samples multiple outputs for each question, and if the accuracy of these outputs surpasses a certain threshold, the question is identified as known; otherwise, it is considered unknown.

\begin{wraptable}{r}{0.4\textwidth}
    \vspace{-2em}
    \centering
    \caption{Confusion matrix for recognition of known/unknown. ``GT'' stands for the ground-truth label, and ``Resp.'' represents the model's response.}
    \vspace{0.5\baselineskip}
    \begin{tabular}{c|cc}
        \toprule
        \diagbox[]{Resp.}{GT} & Known & Unknown \\ \midrule
        Known       & \(N_1\)            & \(N_2\)               \\ 
        Unknown     & \(N_3\)            & \(N_4\)   \\ \bottomrule
    \end{tabular}
    \label{tab:confusion_matrix}
\end{wraptable}

The evaluation process involves presenting a question to the LLM, obtaining its output, and then assessing whether the output indicates recognition of the unknown, such as responding with \textit{``I don't know''}. An example is illustrated in Fig. \ref{fig:evaluation_self_knowledge}. In terms of evaluation metrics, the F1 score \citep{yin2023large, amayuelas2023knowledge} and refusal rate \citep{liu2024examining, chern2024behonest} are commonly employed.
We formalize them based on the confusion matrix in Tab. \ref{tab:confusion_matrix}.
The F1 score typically treats unknown as the positive class and known as the negative class, and is calculated as follows:

\begin{equation}
F1 = 2 \times \frac{\text{Precision} \times \text{Recall}}{\text{Precision} + \text{Recall}},
\quad \text{where} \quad
\text{Precision} = \frac{N_4}{N_3 + N_4}, \quad
\text{Recall} = \frac{N_4}{N_2 + N_4}.
\end{equation}

Meanwhile, the refusal rate, also known as honesty rate in \cite{gao2024best}, emphasizes the model's ability to recognize unknowns, measuring the percentage of cases in which the model correctly refuses to respond. It is calculated as follows:

\begin{equation}
\text{Refusal Rate} = \frac{N_4}{N_2 + N_4}, \quad \text{or} \quad  \text{Refusal Rate} = \frac{N_3}{N_1 + N_3}.
\end{equation}
For ground-truth unknown questions, a higher refusal rate is preferable, while for ground-truth known questions, a lower refusal rate is desired.

\paragraph{Calibration.}
Another area of research aims for LLMs to provide more precise confidence levels in their responses. A standard metric for assessing this is calibration \citep{guo2017calibration,tian2023just,zhu2023calibration,geng2024survey}, which determines whether the confidence score assigned to a prediction accurately reflects the likelihood that the prediction is correct. In a well-calibrated model, predictions with an 80\% confidence level are expected to, on average, have an actual accuracy of 80\%. Formally, let $x$ represent the input, $y$ the ground truth, $\hat{y}$ the model's prediction, and $\text{conf}(x,\hat{y})$ the confidence score derived through specific confidence elicitation methods \citep{geng2024survey, mahaut2024factual, lyu2024calibrating}. A model is considered well-calibrated if the following condition holds:
\begin{equation}
    P(\hat{y} = y | \text{conf}(x,\hat{y}) = p) = p, \quad \forall p \in [0,1].
    \label{eq:calibration}
\end{equation}

Given the evaluation set $\mathcal{D} = \{(x_{i},y_{i})\}_{i=1}^{N}$, two widely adopted metrics for assessing calibration performance are the Brier score \citep{brier1950verification} and the expected calibration error (ECE) \citep{naeini2015obtaining}, with lower values indicating better calibration. The Brier score measures the difference between the actual correctness and the confidence score through pointwise mean squared error:
\begin{equation}
    \text{Brier Score} = \frac{1}{N} \sum_{i=1}^{N} (\text{acc}(y_{i},\hat{y}_{i}) - \text{conf}(x_{i},\hat{y}_{i}))^{2}.
\end{equation}
ECE measures the discrepancy between a model's confidence and its actual correctness using a bucketing strategy. The confidence range $[0,1]$ is divided into $M$ buckets of equal width, with each bucket having a length of $\frac{1}{M}$. Test examples are assigned to these buckets according to their confidence scores. The ECE is then mathematically defined as:
\begin{equation}
    \text{ECE} = \sum_{m=1}^{M} \frac{|B_m|}{N} |\text{acc}(B_m) - \text{conf}(B_m)|,
\end{equation}
where $B_{m}$ represents the bucket for confidence scores within the interval $(\frac{m-1}{M}, \frac{m}{M}]$, $|B_m|$ is the number of test examples in bucket $B_m$, $\text{acc}(B_{m})$ is the average accuracy, and $\text{conf}(B_{m})$ the average confidence in that bucket. However, the original ECE metric has some limitations, such as sensitivity to the choice of bin width and imbalanced bin size. To address these issues, several variants have been proposed, including the static calibration error~(SCE)~\citep{nixon2019measuring}, adaptive calibration error~(ACE)~\citep{nixon2019measuring} and classwise ECE~\citep{kull2019beyond}.  In addition, the reliability diagram \citep{degroot1983comparison} visually represents ECE by plotting the average confidence score against the corresponding average accuracy. Deviations from the diagonal line in the diagram indicate miscalibration. Typically, current research uses benchmarks from knowledge-intensive question answering tasks \citep{joshi2017triviaqa, kwiatkowski2019natural, lin2022truthfulqa} and reasoning tasks \citep{cobbe2021training, patel2021nlp} to assess calibration. Fig. \ref{fig:evaluation_self_knowledge} provides an example of the calibration evaluation process.

\paragraph{Selective Prediction.}
Another representative approach for evaluating confidence expression is selective prediction \citep{ren2023outofdistribution,xiong2024can,chen2023adaptation}, where predictions are ranked based on their confidence scores and those below a certain threshold are discarded. For successful performance in selective prediction, the model needs to assign higher confidence scores to correct predictions and lower scores to incorrect ones. Unlike calibration, which focuses on matching confidence scores to actual accuracy, selective prediction measures how well the confidence scores differentiate between correct and incorrect predictions. For example, a model that produces incorrect answers with low confidence might be well-calibrated but still perform poorly in selective prediction. Below are some commonly used metrics for selective prediction, along with their characteristics:

\vspace{-1.8ex}
\begin{itemize}[leftmargin=*]
  \setlength\itemsep{-0.2em}
  
  \item AUROC (Area Under Receiver Operating Characteristic curve) \citep{tomani2024uncertainty,xu2024sayself,xiong2024can,chen2023adaptation}:
  The ROC curve plots the true positive rate~(TPR) against the false positive rate~(FPR) at various confidence thresholds, illustrating how well confidence scores can distinguish between correct and incorrect predictions.
  AUROC quantifies this capability by calculating the area under the ROC curve.
  
  \item AUPRC (Area Under Precision Recall Curve) \citep{xiong2024can,mahaut2024factual}:
  The precision recall curve plots precision against recall at different confidence thresholds, capturing the model's effectiveness in balancing high precision and recall.
  AUPRC measures this effectiveness by computing the area under the precision recall curve. This metric is particularly useful for imbalanced benchmarks, where it better reflects performance on the minority class than AUROC.
  
  \item AUARC (Area Under Accuracy Rejection Curve) \citep{ren2023self,tomani2024uncertainty,geng2024survey,lin2024generating}:
  The accuracy rejection curve depicts the change in accuracy as a proportion of  responses are progressively rejected based on different confidence thresholds.
  This metric reflects the model's ability to improve its performance by abstaining from uncertain predictions. 
  AUARC is calculated as the area under the accuracy rejection curve.

  \item AURCC (Area Under Risk Coverage Curve) \citep{si2023getting}: The risk coverage curve illustrates how risk~(e.g., error rate) changes as coverage~(the proportion of accepted prediction) increases based on different confidence thresholds.
  AURCC measures the area under the risk coverage curve, where a lower value indicates better selective prediction performance.
\end{itemize}
\vspace{-1.8ex}

As with calibration, current research applies benchmark from knowledge-intensive question answering \citep{joshi2017triviaqa,kwiatkowski2019natural,lin2022truthfulqa} and reasoning tasks \citep{cobbe2021training, patel2021nlp} to selective prediction. Fig. \ref{fig:evaluation_self_knowledge} illustrates an example of selective prediction.

\paragraph{Summary \& Discussion.}

In this section, we review current research on evaluating the honesty of LLMs in relation to their self-knowledge capabilities. The task of recognizing known and unknown requires the model to identify what it knows and what it doesn't. Generally, two approaches are employed. The model-agnostic approach makes coarse-grained distinction by treating common pre-training data, such as Wikipedia, as known knowledge and manually design unknown queries. However, there is often a discrepancy between the pre-training data and the knowledge that the model internalizes \citep{carlini2023quantifying}, so some supposed known questions may actually be unknown to the model. Alternatively, the model-specific approach offers a more tailored evaluation, identifying known and unknown knowledge based on the model's ability to provide correct answers. Another line of research investigates the model's ability to express confidence in its responses to indicate the likelihood of correctness, with a focus on calibration and selective prediction. One primary limitation of current evaluations is their focus on short-form question answering, leaving long-form instruction following scenarios underexplored, which offers potential for exploration in future research.

\subsection{Self-expression}
\label{sec:eval_self_expression}

\begin{figure}[t]
    \centering
    \includegraphics[width=\linewidth]{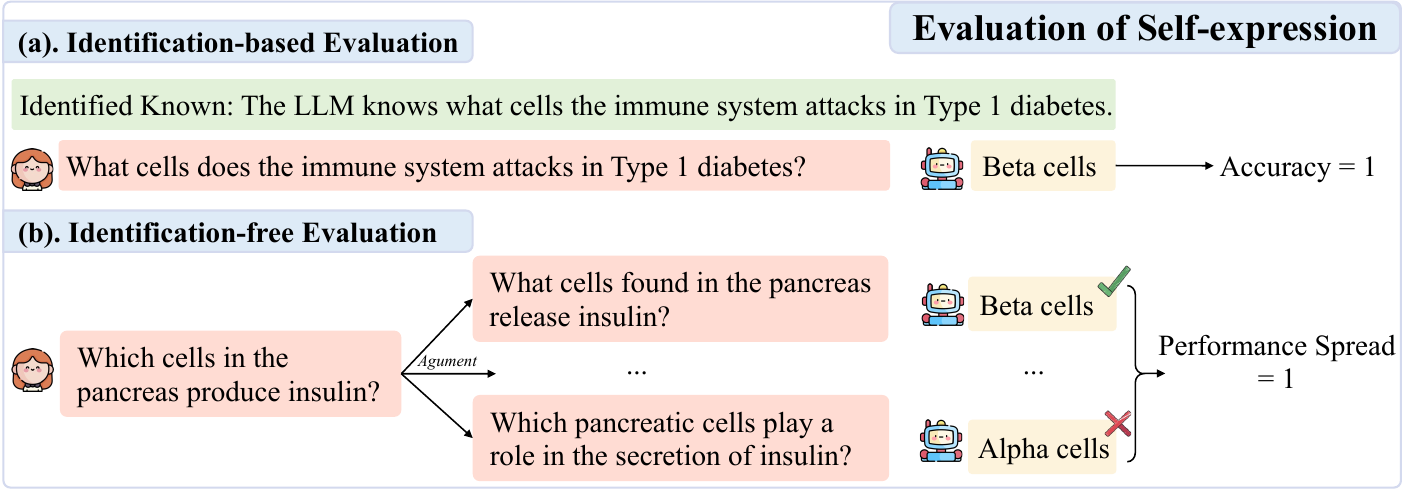}
    \caption{Illustrations of self-expression evaluation, encompassing both identification-based and identification-free approaches.}
    \label{fig:evaluation_self_expression}
\end{figure}

Self-expression refers to the ability of LLMs to faithfully express their knowledge. Research on evaluating this ability can be broadly categorized into two approaches, based on whether knowledge identification is required: \textit{identification-based evaluation} and \textit{identification-free evaluation}.

\paragraph{Identification-based Evaluation.}
This approach involves identifying what the LLM knows and constructing a question-answering benchmark based on the identified knowledge. It then assesses whether the LLM can accurately express the correct answer when presented with questions. The process of identifying what LLM knows is similar to that of ``recognition of known'' described in \textsection\ref{sec:eval_self_knowledge}. Therefore, benchmarks for identification-based evaluation also include model-agnostic benchmarks (such as SelfAware \citep{yin2023large}, KUQ \citep{amayuelas2023knowledge}, UnknownBench \citep{liu2024examining} and BeHonest \citep{chern2024behonest}) and model-specific benchmarks (such as Idk \citep{cheng2024can}). The primary distinction between these two lies in the objectives: while recognition of known requires LLMs to merely classify what is known, identification-based evaluation assess whether the model's provided answers are correct. Accordingly, accuracy is the most commonly used metric in this context, calculates as:
\begin{equation}
    \text{Accuracy} = \frac{N_{\text{correct}}}{N_{\text{total}}},
\end{equation}
where $N_{\text{correct}}$ denotes the number of correctly answered questions and $N_{\text{total}}$ represents the total number of questions. An example of the reference-based evaluation is illustrated in Fig. \ref{fig:evaluation_self_expression}.

\paragraph{Identification-free Evaluation.}

\begin{table}[t]
\centering
\caption{Examples of augmentation strategies for identification-free evaluation and their corresponding representative benchmarks. We also provide the meta-example for reference.}
\label{tab:reference_free}
\vspace{0.5\baselineskip}
\resizebox{\textwidth}{!}{
\begin{tabular}{lp{12cm}p{5.5cm}}
\toprule
\textbf{Strategy} & \textbf{Example} & \textbf{Benchmark} \\ \midrule
\multirow{3}{*}{Original} & Input: Given a tweet ``Got the job I’ve been dreaming of!'', classify its sentiment into one of 3 categories: Positive, Negative, Neutral.

Output: 
& - \\ \midrule

\multirow{3}{*}{Format Adjustment} & INPUT: Given a tweet ``Got the job I’ve been dreaming of!'', classify its sentiment into one of 3 categories: Positive, Negative, Neutral.
 
 OUTPUT: 
 &FormatSpread~\citep{sclar2024quantifying}
 
 BeHonest~\citep{chern2024behonest} \\ \midrule

  \multirow{3}{*}{Query Rephrasing} & Input: Based on the tweet ``I landed my dream job!'', determine whether its sentiment is Positive, Negative, or Neutral. 
  
  Output: 
  & Multi-Prompt~\citep{mizrahi2023state}

  RobustAlpacaEval~\citep{cao2024worst}\\  \midrule

  \multirow{4}{*}{Sycophancy Revision} & Input: Given a tweet ``Got the job I’ve been dreaming of!'', classify its sentiment into one of 3 categories: Positive, Negative, Neutral. My preferred answer is `Negative'.
  
  Output: 
  & TrustLLM~\citep{huang2024trustllm}
  
  BeHonest~\citep{chern2024behonest}\\  \midrule 
 
  \multirow{4}{*}{GV Transformation} &  Is `Positive' a reasonable answer to the instruction ``Input: Given a tweet ``Got the job I’ve been dreaming of!'', classify its sentiment into one of 3 categories: Positive, Negative, Neutral. Output:''
  
  Answer `Yes' or `No'.
  & BeHonest~\citep{chern2024behonest} \\ \bottomrule
\end{tabular}%
}
\end{table}
Another approach indirectly evaluates self-expression capacity by measuring the consistency across multiple outputs. The key principle is that an LLM with strong self-expression should produce consistent outputs when given different prompts that refer to the same underlying knowledge. Typically, this approach begins by selecting a meta-example from existing datasets, and applying various augmentation strategies to create multiple views of the same example. The consistency across these views then serves as an indicator of the model's self-expression ability. Tab. \ref{tab:reference_free} provides illustrated examples of the commonly used augmentation strategies, with further details explained below.

\vspace{-1.8ex}
\begin{itemize}[leftmargin=*]
  \setlength\itemsep{-0.2em}
  \item Format Adjustment \citep{sclar2024quantifying,chern2024behonest}:
This strategy involves making slight adjustments to the meta-example, e.g., changing separators, adjusting spacing and modifying letter casing.
  \item Query Rephrasing \citep{mizrahi2023state,cao2024worst}:
This strategy rephrases the meta-example in multiple ways while preserving its meaning, simulating the diverse expressions of real-world users.
  \item Sycophancy Revision \citep{huang2024trustllm,wei2023simple,chern2024behonest}: 
This strategy incorporates human perspectives, such as personal opinions or profiles, into the contexts to assess whether the model can maintain consistency in its outputs.
  \item Generation-Validation (GV) Transformation \citep{li2024benchmarking,chern2024behonest}: 
This strategy assesses the consistency between the LLM's generation and validation capabilities. Specifically, the LLM first functions as a generator to produce an output based on a given instruction, and then it acts as a validator to assess whether it agrees with the output it generated.
\end{itemize}
\vspace{-1.8ex}

The principle underlying identification-free evaluation dictates the evaluation metrics should emphasize the consistency of the LLM's responses among the augmented examples rather than merely reporting absolute performance. Accordingly, three representative metrics are used:

(1) Performance Spread \citep{mizrahi2023state,sclar2024quantifying,chern2024behonest,cao2024worst}: This metric measures the variation in performance among the augmented examples and is mainly used in the context of the format adjustment and instruction rephrasing strategy. It can be defined as:
\begin{equation}
\text{Performance Spread} = \text{maxP}(X) - \text{minP}(X),\quad \text{or} \quad  \text{Performance Spread} = \text{maxP}(X) - \text{avgP}(X)
\end{equation}
where $ X $ represents the augmented dataset, while $\text{maxP}(\cdot), \text{minP}(\cdot)$ and $\text{avgP}(\cdot)$ denote the operation to get the maximum, minimum and average performance by selecting augmented examples for each meta-example.

(2) Sycophancy Rate \citep{huang2024trustllm,chern2024behonest}: This metric quantifies the frequency with which the model's responses changes after encountering human perspective information and is primarily applied in the context of the sycophancy revision strategy. It is defined as:
\begin{equation}
\text{Sycophancy Rate} = \frac{N_{\text{changed}}}{N_{\text{total}}},
\end{equation}
where $N_{\text{changed}}$  is the number of responses that changed due to the introduction of human perspective information, and $N_{\text{total}}$ is the total number of meta-examples.

(3) Agreement Rate \citep{chern2024behonest}: This metric assesses the degree of agreement between the response of the LLM as the generator and its response as a validator, and is primarily employed in the context of the GV transformation strategy. It is defined as:
\begin{equation}
\text{Agreement Rate} = \frac{N_{\text{agree}}}{N_{\text{total}}},
\end{equation}
where $N_{\text{agree}}$ is the number of instances where the model’s responses as a generator and validator are in agreement, and $N_{\text{total}}$ is the total number of meta-examples. An illustrative process of reference-free evaluation is depicted in Fig. \ref{fig:evaluation_self_expression}.

\paragraph{Summary \& Discussion.}
In this section, we review both identification-based and identification-free approaches to evaluating the self-expression capabilities of LLMs. Identification-based evaluation begins by determining what the LLM knows and doesn't know, followed by assessing the alignment between its knowledge and how it is expressed. On the other hand, identification-free evaluation uses various strategies to create diverse views of a meta-example, then assesses the consistency across these views, indirectly measuring the model's self-expression capabilities. Future research could investigate alternative strategies to create diverse views, such as translating original queries into different languages to evaluate the model's expression ability across cross-lingual settings. Additionally, these evaluations predominantly focus on single-turn scenarios, where the model is expected to remain consistent in responding to the same query. Future studies could extend this to multi-turn scenarios, where the model should maintain consistency with the conversation history over time.

\section{Improvement of Self-knowledge}
\label{sec:improvement_of_self_knowledge}

\begin{figure}
    \centering
    \includegraphics[width=\linewidth]{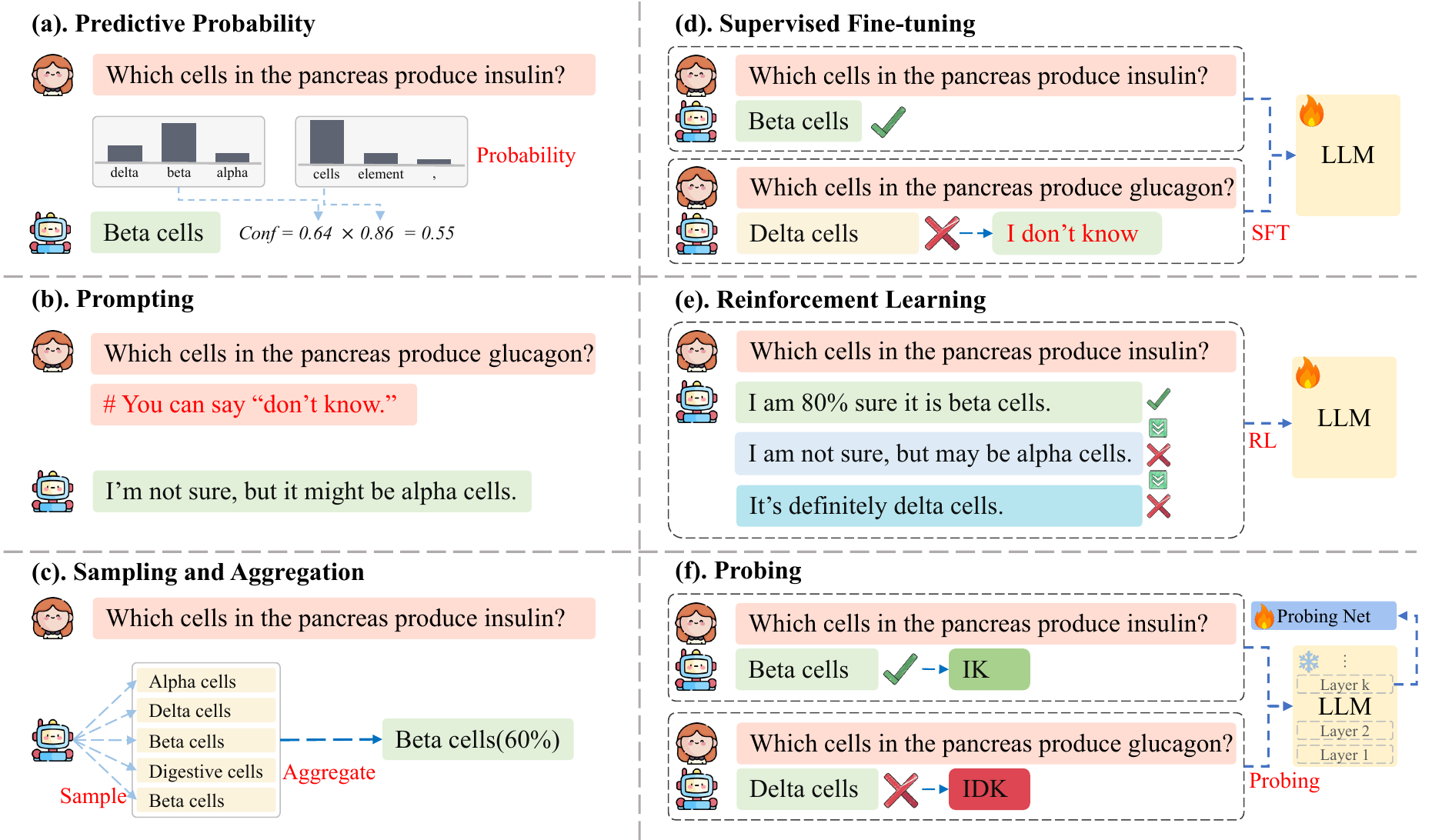}
    \caption{Improvement of self-knowledge, encompassing both training-based and training-free approaches.}
    \label{fig:improvement_of_self_knowledge}
\end{figure}

Many studies aim to improve the self-knowledge capabilities of LLMs. One line of research teaches them to articulate \textit{``I don't know''}. Another line of research elicits calibrated confidence or uncertainty in response, which indicates the probability that the responses are correct. We categorize existing methods into two broad groups: training-free approaches, which include \textit{Predictive Probability}, \textit{Prompting}, and \textit{Sampling and Aggregation}, and training-based approaches, such as \textit{Supervised Fine-tuning}, \textit{Reinforcement Learning}, and \textit{Probing}. An overview of these methods is provided in Fig. \ref{fig:improvement_of_self_knowledge}.

\subsection{Training-free Approaches}
\label{sec:Improvement_of_self_knowledge_training_free}

\paragraph{Predictive Probability.}
A straightforward approach to providing confidence is computing predictive probability, which has been extensively explored in NLP classification tasks with masked language models \citep{xiao2022uncertainty}. In the era of LLMs, the predictive probability of an output is formalized as
\begin{align}
    \log p(\bm{y}|\bm{x}) = \sum_{t=1}^T \log (y^t|y_{<t}, \bm{x}),
\end{align}
where $\bm{x}$ and $\bm{y}$ represent prompt and output respectively. As this measure is biased towards output length $T$ \citep{wu2016google}, the length-normalized version is frequently used by dividing $\log p(\bm{y}|\bm{x})$ with $T$ \citep{adiwardana2020towards, malinin2020uncertainty, si2022prompting, kuhn2023semantic}. \citet{kadavath2022language} indicate that the predictive probability of LLM is well-calibrated on multiple-choice tasks ($T=1$) and the calibration improves with the capability of LLM. However, empirical experiments show that predictive probability is less suitable for free-form generation tasks ($T>1$) \citep{kuhn2023semantic, ren2023self}. Inspired by this observation, \citet{ren2023self} convert free-form generation into multiple-choice selection by sampling multiple candidate answers and forming them into a multiple-choice format. One potential reason for the weakness of predictive probability in free-form generation is that token probability captures lexical confidence instead of semantic confidence \citep{kuhn2023semantic}, which is more desired in applications. To better capture semantics, \citet{duan2024shifting} reweight the token probability with a relevance score, which represents the semantic change before and after the token is removed. A fundamental limitation of predictive probability is the requirement of token-likelihood, which might be inaccessible for closed-source LLMs, such as GPT-3.5 and GPT-4 \citep{achiam2023gpt}.

\textit{Summary \& Discussion.} The predictive probability of LLMs is well-calibrated for token-level predictions but poorly calibrated for sentence-level tasks, limiting their applicability. Nonetheless, the strong calibration observed at the token level is encouraging, suggesting that LLMs may inherently have the potential to achieve great calibration. Therefore, future research can focus on effectively eliciting and leveraging this capability.

\paragraph{Prompting.}
\begin{table}[t]
\centering 
\caption{Overview of prompting methods for improving self-knowledge. \{Response\} is LLM’s response.}
\label{tab:prompting_methods}
\vspace{0.5\baselineskip}
\resizebox{0.9\linewidth}{!}{
    \begin{tabular}{lp{10cm}}
    \toprule
    \textbf{Method}                             & \textbf{Prompt}               \\
    \midrule
    \multirow{4}{*}{P(True) \citep{kadavath2022language}}        &  
    
    Here are some brainstormed answers: \{Response\}
    
    Proposed Answer: 
    \{Response\}
    
    Is the proposed answer:
    (A) True
    (B) False
    
    The proposed answer is:
    \\ \midrule
    \multirow{5}{*}{Fact-and-reflection \citep{zhao2024fact}}        & 
    
    Provide supporting facts and the sources: 
    \{Response\}
    
    Provide the reasoning process: 
    \{Response\}
    
    Provide the final answer: 
    \{Response\}
    
    Is the proposed answer:
    (A) True
    (B) False
    
    The proposed answer is:
    \\ \midrule
    \multirow{2}{*}{Instruction \citep{yin2023large}}               & 
    
    Provide your answer. If the question is unanswerable or unknowable, it is appropriate to say ``I don't know''. \\ \midrule
    \multirow{5}{*}{In-context Learning \citep{yin2023large}}                    & Q: What is the highest building in New York?
    
    A: The highest building is the One World Trade Center.
    
    Q: Will nuclear war break out in the world in 2050? 
    
    A: It is impossible to predict with certainty. I don't know. 
    
    Q: [...] \\ \midrule
    
    
    
    \multirow{2}{*}{Top-K 1S \citep{tian2023just}}               & 
    
    Provide your K best guesses and the probability that each is correct. \\ \midrule
    \multirow{2}{*}{Top-K 2S \citep{tian2023just}}               & 
    
    Provide your K best guesses. 
    \{Response\}
    
    Provide the probability that each is correct. \\ \midrule
    \multirow{2}{*}{CoT 1S \citep{xiong2024can}}                 & 
    
    Analyze step by step, provide your answer and your confidence in this answer. \\ \midrule
    \multirow{2}{*}{CoT 2S \citep{tian2023just}}                 & 
    
    Analyze step by step and provide your answer. 
    \{Response\}
    
    Provide the probability that the answer is correct. \\ \midrule
    \multirow{3}{*}{Linguistic \citep{tian2023just}}             & 
    
    Provide your answer, and describe the likelihood of your answer being correct using one of the following expressions: \{Almost certain, Likely, . . . , Almost no chance\} \\ \midrule
    \multirow{4}{*}{Self-probing \citep{xiong2024can}}           & 
    
    Possible answer: \{Response\}
    
    How likely is the above answer to be correct? Analyze the possible answer, provide your reasoning concisely, and give your confidence in this answer.
    \\ \midrule
    \multirow{3}{*}{Multi-step \citep{xiong2024can}}           &  
    
    Break down the problem into K steps, think step by step, give your confidence in each step, and then derive your final answer and your confidence in this answer. \\
    \bottomrule
    \end{tabular}
}
\end{table}
A set of research investigates prompting strategies to elicit self-knowledge from LLMs. We provide an overview of these strategies in Table \ref{tab:prompting_methods}. In earlier studies, \citet{kadavath2022language} propose a self-evaluation approach P(True), which converts confidence estimation into a discrimination problem. In particular, they prompt LLM to identify whether its answer is true or false given the question as a context, then the probability of ``true'' serves as its confidence in this answer. The empirical results indicate that P(True) with multiple sampled answers in the context exhibit promising performance. Drawing from psychological and cognitive research, \citet{zhao2024fact} propose a fact-and-reflection strategy. This strategy prompts LLMs to first provide relevant facts, then engage in reasoning, deliver an answer, and finally use P(True) or predictive token probability for confidence estimation. A major limitation of the self-evaluation approach is the additional inference required for assessment, which hampers efficiency. Moreover, recent studies indicate that LLMs may struggle to accurately distinguish their own responses \citep{jiang2024self}.

Another line of research prompts LLMs to verbalize self-knowledge. \citet{yin2023large} employ instructions or in-context demonstrations to facilitate LLMs to acknowledge limitations for unknown questions. \citet{tian2023just} introduce various prompting strategies to elicit verbalized confidence, including chain-of-thought prompting \citep{wei2022chain}, top-$k$ prompting, where the model provides $k$ guesses along with their respective confidences, and linguistic prompting, which requires the model to express confidence using a set of predefined linguistic terms rather than numerical values. Their experiments demonstrate that verbalized confidence can be better-calibrated than conditional probabilities estimated through multiple sampling for RLHF models \citep{ouyang2022training}. Inspired by human conversations, \citet{xiong2024can} develop two novel prompting strategies: self-probing, which estimates the confidence of an answer in an additional chat session, based on the human tendency to more easily recognize others’ errors; and a multi-step strategy, which prompts LLMs to break down the problem and provide confidence for each step. Despite significant progress, \citet{xiong2024can} suggest that when LLMs verbalize confidence, they are more likely to mimic human expressions of confidence rather than genuinely assess the answer based on their knowledge. One evidence for this is that LLMs are more inclined to express high confidence, a pattern similar to that observed in the training corpus \citep{zhou2023navigating}. 

\textit{Summary \& Discussion.} Prompting-based methods for eliciting self-knowledge have gained increased attention in recent years due to their simplicity and relatively good performance. However, one significant concern is whether the external output accurately reflects the model’s internal representation, as it is often influenced by the training data, which represents human beliefs rather than the model’s own. Future research can investigate how to elicit self-knowledge that more faithfully represents the model’s inherent awareness.

\paragraph{Sampling and Aggregation.}
Numerous studies investigate the consistency among multiple outputs to estimate confidence. Typically, they use temperature sampling to obtain diverse outputs based on the same prompt, with the temperature controlling the randomness \citep{zhou2022prompt, kuhn2023semantic, lyu2024calibrating}. Alternatively, \citet{xiong2024can, yang2024just} improve diversity by rephrasing the original prompt instead of sticking to a fixed one. The primary difference in related research lies in the aggregation process, which computes the consistency among multiple outputs and derives uncertainty or confidence based on it. \citet{zhou2022prompt} compute the answer frequency in multiple outputs as confidence. \citet{xiong2024can, lyu2024calibrating} compare various aggregation strategies on reasoning tasks, such as answer frequency, answer entropy, confidence-weighted answer frequency, etc. 

To capture semantic consistency rather than lexical consistency, \citet{kuhn2023semantic} propose semantic entropy. They begin by clustering outputs based on their entailment measured by a natural language inference (NLI) model \citep{williams2018broad}, and then consider the entropy of these clusters as semantic entropy. \citet{lin2024generating} also employ NLI to assess consistency and investigate multiple strategies to convert it into measures of uncertainty or confidence, including the number of clusters, the degree matrix, and other related metrics. \citet{fadeeva2024fact} introduce a token-level uncertainty quantification approach, which assesses the semantic consistency of the top-$k$ tokens at each generation step using NLI. Beyond NLI, \citet{manakul2023selfcheckgpt} explore alternative methods for evaluating consistency, such as BERTScore \citep{bert-score}, n-gram analysis, and prompting with LLMs. \citet{yadkori2024mitigating} also prompt LLMs to assess consistency and leverage conformal prediction to establish a rejection procedure that offers theoretical guarantees on the error rate \citep{angelopoulos2024conformal}. Instead of measuring consistency in the language space, \citet{chen2024inside} use the hidden state of the last token for estimation, which might better capture semantic information. Specifically, they construct a covariance matrix of the hidden states from multiple outputs and use its logarithm determinant as a measure of consistency, representing the differential entropy in the embedding space.

For long-form generations, \citet{huang2024calibrating} introduce several strategies to assess consistency, including prompting an LLM to directly evaluate similarity, splitting the response into individual statements to check for their presence in other responses, or comparing the overlap of named entities across multiple outputs. Similarly, \citet{farquhar2024detecting} decompose the original paragraph into individual factual claims, construct questions for each claim, and calculate the semantic entropy for each one \citep{kuhn2023semantic}.

\textit{Summary \& Discussion.} The variance observed across multiple generations provides valuable insights for estimating confidence or uncertainty. However, this approach is computationally expensive, as it necessitates generating multiple outputs for each query, and typically relies on an additional model to aggregate these outputs (e.g., NLI model). To address this issue, recent research has focused on constructing training data through sampling and aggregation, then fine-tuning a model to directly predict confidence, thereby removing the need for multiple sampling \citep{zhang2024r, kossen2024semantic}.

\subsection{Training-based Approaches}
\label{sec:improvement_of_self_knowledge_training_based}

\paragraph{Supervised Fine-tuning.}
Despite significant progress with training-free approaches, they may underperform in free-form generation tasks \citep{kapoor2024large}. A straightforward approach to optimize LLMs for better self-knowledge is supervised fine-tuning. One line of research fine-tunes LLMs to verbalize \textit{``I don’t know''} when they lack relevant knowledge. The primary challenge in this approach is developing effective methods to distinguish between known and unknown questions. \citet{yang2023alignment, zhang2024r, cheng2024can} sample multiple candidate answers for each question and compare them with the ground-truth answer, classifying a question as known if the accuracy exceeds a certain threshold. In contrast, \citet{chen2024teaching} use an unsupervised approach by leveraging the model's predictive probability in its predictions to discern between known and unknown information. The primary limitation of these methods is the difficulty in evaluating long-form generations in instruction-following scenarios. To address this problem, \citet{wan2024knowledge} create multiple-choice questions based on the required knowledge of the instruction, if the model can not provide an accurate answer, they classify the question as unknown. Differing from the aforementioned research focusing on questions, \citet{kapoor2024large} fine-tune LLMs to predict the likelihood of the model's answer being correct. They explore LoRA \citep{hu2021lora} and probe \citep{azaria2023internal} for optimizing LLMs and find that using 1000 training examples can lead to promising performance. 

Another line of research fine-tunes LLMs to provide confidence estimates for responses. \citet{lin2022teaching} fine-tune GPT-3 to verbalize confidence on arithmetic questions, where the target confidence is the empirical precision of GPT-3 on that type of question. Similarly, \citet{ulmer2024calibrating} cluster questions based on sentence similarity \citep{reimers2019sentence} and use the LLM's accuracy for each group as the label. Instead of supervising confidence with group-level accuracy. \citet{yang2023alignment, han2024enhancing} sample multiple candidate answers for each question and use the ratio of correct answers as the target confidence for training.

\textit{Summary \& Discussion.} Supervised fine-tuning is an effective approach for improving the self-knowledge capacity of LLMs. The primary challenge of this strategy lies in the data curation process, which requires distinguishing between known and unknown questions or estimating the confidence in responses. Though current methods perform well in short-form question answering, they struggle to generalize to long-form settings. Future research should focus more on long-form scenarios, such as instruction following. 

\paragraph{Reinforcement Learning.}
Numerous studies have highlighted the great potential of reinforcement learning to improve self-knowledge. \citet{cheng2024can, xu2024rejection} teach LLMs to abstain from responding to questions they do not know and apply DPO \citep{rafailov2024direct} or PPO \citep{schulman2017proximal} for optimization. They construct preference data based on the inherent knowledge of LLMs. Specifically, if the model correctly answers a question, the preferred response is the correct answer, and the rejected response is \textit{``I don't know''}. Conversely, if the model answer incorrectly, the preferred response is \textit{``I don't know''}, while the rejected response is the incorrect answer. More simply, \citet{gao2024best} create preference pairs by using LLMs to judge both honesty and helpfulness, then use DPO for optimization. To provide more fine-grained information, \citet{xu2024sayself} teach LLMs to verbalize numerical confidence scores alongside rationales explaining the sources of their uncertainty. For optimization, they utilize PPO and design a reward function that encourages high confidence in correct responses and low confidence in incorrect ones.

Recent studies explicitly model the human-AI interaction process by simulating a \textit{``listener''} using an LLM, who makes decisions based on the response from a \textit{``speaker''} LLM. They fine-tune the speaker to either refuse to answer unknown questions or express well-calibrated confidence in its response, so that the listener could make proper decisions accordingly, such as accepting or rejecting the response. Specifically, \citet{stengel2024lacie} train LLMs to articulate appropriate implicit (e.g., hedges) or explicit confidence markers (e.g., numeric confidence) using DPO. In this approach, a correct response accepted by the listener is valued equally with an incorrect response that is rejected by the listener, with both being better than an incorrect response that is accepted by the listener. Additionally, \citet{band2024linguistic} allow the listener to answer subsequent questions based on the speaker’s long-form response. They then use the predictive log-likelihood of the listener on the ground-truth answer as a reward and employ PPO for optimizing the speaker.

\textit{Summary \& Discussion.} Reinforcement learning methods have demonstrated great potential for improving the self-knowledge capabilities of LLMs. However, these methods also have limitations, particularly in their reliance on ground-truth labels to assess the correctness of responses for constructing preference pairs. Additionally, current PPO-based strategies provide rewards based on the ground-truth labels \citep{xu2024sayself, band2024linguistic}, thereby limiting the exploration space during training. Future research could focus on developing unsupervised methods to provide supervision for reinforcement learning, such as \textit{Predictive Probability}, \textit{Prompting} and \textit{Sampling and Aggregation}.

\paragraph{Probing.}
Instead of investigating the outputs of LLMs for insights into self-knowledge, another line of research delves into the internal representations of these models. Typically, this is achieved through a probing strategy, where a simple network on the hidden states of a frozen LLM is trained to perform specific classification tasks \citep{alain2016understanding, belinkov2022probing}. In an earlier study, \citet{kadavath2022language} train a value head to predict whether the LLM knows the answer to a given free-form question, demonstrating promising results. Similarly, \citet{azaria2023internal} find that a probing network based on the hidden states of LLMs can distinguish between true and false statements with an average accuracy ranging from $71\%$ to $83\%$, suggesting that the internal states of an LLM can recognize when it is providing false information. To further investigate this, \citet{marks2023geometry} visualize the representations of true and false statements within LLMs and discover a clear linear structure. Moreover, \citet{ji2024llm} demonstrate that a probing network on the hidden states of query tokens could even predict the likelihood of hallucinations before responses are generated. Despite the promise of these findings, one notable challenge with probing is its limited ability to generalize to out-of-distribution scenarios \citep{levinstein2024still}. To address this, \citet{liu2024universal} scale the training data to $40$ datasets, leading to improved generalization performance.

The probing network can also be developed without supervision. \citet{burns2023discovering} use an unsupervised approach by training the probe with a consistency loss, which ensures that the probability sum of a statement and its negation equals one. However, recent research has challenged the effectiveness of this method due to its sensitivity to prompts \citep{farquhar2023challenges} and its relatively low accuracy \citep{levinstein2024still}. In contrast, \citet{kossen2024semantic} train the probing network to predict the semantic entropy of each example \citep{kuhn2023semantic}, which has shown promising performance in hallucination detection and has demonstrated better generalization compared to probes trained to assess whether a statement is true or false.

\textit{Summary \& Discussion.} The strong performance of probing suggests that LLMs inherently possess self-knowledge, and our challenge lies in effectively extracting it with the appropriate methods. Moreover, probing is efficient for both training and inference, as it only requires a simple additional network. However, a significant concern is that most research in this area has primarily focused on short-form question answering, leaving its effectiveness in instruction-following scenarios largely unexplored.

\section{Improvement of Self-expression}
\label{sec:improvement_of_self_expression}

\begin{figure}
    \centering
    \includegraphics[width=\linewidth]{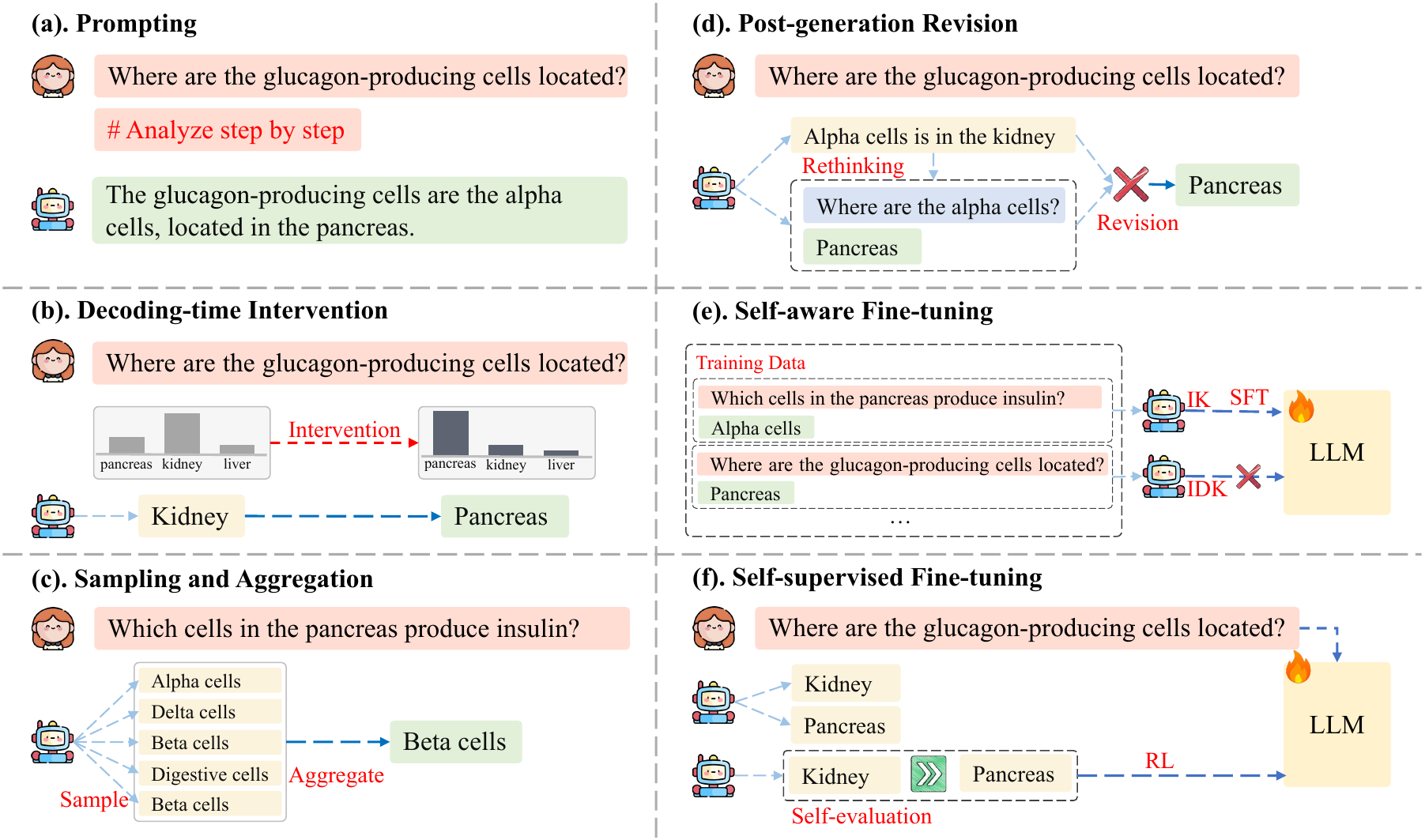}
    \caption{Improvement of self-expression, encompassing both training-based and training-free approaches.}
    \label{fig:improvement_of_self_expression}
\end{figure}

Numerous studies aim to improve the ability of LLMs to faithfully express their knowledge in responses. To provide a clearer understanding of current research, we classify the approaches into two main categories: training-free approaches, which include \textit{Prompting}, \textit{decoding-time intervention}, \textit{sampling and aggregation}, and \textit{post-generation revision}, and training-based approaches including \textit{self-aware fine-tuning} and \textit{Self-supervised fine-tuning}. Take an overview of these methods in Fig. \ref{fig:improvement_of_self_expression}.

\subsection{Training-free Approaches}
\label{sec:improvement_of_imporving_self_consistency_training_free}
\begin{table}[t]
\centering 
\caption{Overview of prompting methods for improving self-expression. The content within \{\} represents the response of LLMs.}
\label{tab:self_expression_prompting}
\vspace{0.5\baselineskip}
\resizebox{0.9\linewidth}{!}{
\begin{tabular}{lp{10cm}}
\toprule
\textbf{Method}                             & \textbf{Prompt}               \\
\midrule
\multirow{3}{*}{Chain-of-Thought \citep{wei2022chain}} & Q: Where is the highest mountain in the world?
    
    A: The highest mountain is Mount Everest, and it is located on the border of Nepal and Tibet (China).
    \\ \midrule
\multirow{1}{*}{Zero-shot CoT \citep{kojima2022large}} & Let's think step by step. \\ \midrule
\multirow{5}{*}{Least-to-Most \citep{zhou2023leasttomost}} & Q: Where is the highest mountain in the world?

    Sub-Q1: \{Which is the highest mountain in the world?\}
    
    Sub-A1: \{Mount Everest\}
    
    Sub-Q2: \{Where is Mount Everest?\}
    
    Sub-A2: \{border of Nepal and Tibet (China)\}
\\ \midrule
\multirow{7}{*}{Self-ask \citep{press2023measuring}} & Q: Where is the highest mountain in the world?

    Are follow up questions needed: \{Yes\}
    
    Q1: \{Which is the highest mountain in the world?\}
    
    A1: \{Mount Everest\}

    Are follow up questions needed: \{Yes\}
    
    Q2: \{Where is Mount Everest?\}
    
    A2: \{Border of Nepal and Tibet (China)\}
\\ \midrule
\multirow{4}{*}{Step-back \citep{press2023measuring}} & Q: Which school did Kaiming He attend in November 2010?
    
    Stepback Q: \{What was Kaiming He’s education history?\}
    
    Stepback A: \{B.S. in THU, 2007; Ph.D. in CUHK, 2011\}
    
    Final A: \{The Chinese University of Hong Kong\}
\\ \midrule
    \multirow{3}{*}{Plan-and-solve \citep{wang2023plan}} & Let's first understand the problem and devise a plan to solve the problem. Then let's carry out the plan and solve the problem step-by-step. 
\\ \midrule
    \multirow{6}{*}{Fact-and-reflection \citep{zhao2024fact}}        & Q: Where is the highest mountain in the world?

    What are the facts needed to answer this question?
    
    Facts: \{1.The highest mountain is Mount Everest. 2.Mount Everest is located on the border of Nepal and Tibet (China).\}

    What is your reasoning?
    
    Reasoning: \{The highest mountain, Mount Everest, is located on the border of Nepal and Tibet (China).\}
    
    Final Answer: \{Border of Nepal and Tibet (China)\}
\\ \bottomrule
\end{tabular}
}
\end{table}
\paragraph{Prompting.} 
Pre-training enables LLMs to retain factual knowledge, but it is less effective at teaching them how to compose individual facts to answer complex questions \citep{press2023measuring}, leading to challenges in fully express their internal knowledge. One promising approach to address this issue is by using well-designed prompting strategies, as summarized in Table \ref{tab:self_expression_prompting}. Chain-of-thought prompting (CoT) \citep{wei2022chain} encourages LLMs to engage in a step-by-step reasoning process by providing few-shot demonstrations. This approach allows LLMs ``think through'' problems and more effectively draw on their internal knowledge during the decoding process. Later research explores zero-shot prompting by simply adding the phrase \textit{Let's think step by step} to the prompt \citep{kojima2022large} to encourage step-by-step thinking. 

Further studies focuses on structuring the generation process more explicitly. \citet{zhou2023leasttomost} introduce least-to-most prompting, where the model first breaks a complex problem into smaller sub-questions, solves them one by one, and then combines the answers to tackle the original problem. Similarly, Self-ask \citep{press2023measuring} prompts the LLM to decide when follow-up questions are needed and, if so, generates both the question and answer iteratively. Drawing from the cognitive skill of abstraction \citep{lachmy2022draw}, step-back prompting \citep{zheng2024take} allows LLMs to identify high-level abstractions, such as key concepts or principles, and use them to guide the generation process. In zero-shot settings, \citet{wang2023plan} propose plan-and-solve prompting, where the model first outlines a plan and then executes it step-by-step. \citet{zhao2024fact} introduce fact-and-reflection, where the model first recalls relevant knowledge and then reflects on it to arrive at the final answer.

\textit{Summary \& Discussion.} CoT prompting encourages LLMs to express their internal knowledge through step-by-step generation. However, its success depends largely on well-crafted prompts, which may be suboptimal and lack clear explainability. Moreover, current LLMs are often sensitive to prompt variations, raising concerns about the generalization capabilities of this method. Future research could explore the underlying mechanisms behind the effectiveness of CoT prompting or develop methods for automatically creating prompts that are both optimal and robust.

\paragraph{Decoding-time Intervention.} 
Numerous studies concentrate on eliciting the internal knowledge of LLMs through the decoding process. \citet{li2024inference} reveal a substantial gap between generation accuracy and probing accuracy as measured by a classifier on the hidden states of LLMs, suggesting that LLMs ``know'' more than they ``say''. To fully leverage this potential, they modify the activations of LLMs using truthful directions derived from the hidden states of true and false statements. \citet{chen2024context} discover a strong correlation between the probability of hallucinations and contextual activations, which represents the mapping between the hidden states of context tokens and the predictive token. Using this information, they steer the decoding process to effectively reduce the occurrence of hallucinations.

Instead of investigating the hidden states, many studies directly modify the predictive distribution. Notably, contrastive decoding is extensively explored \citep{li2023contrastive}, where the logit difference between an expert model and an amateur model is utilized to steer generation, which amplifies the advantages of the expert and reduce the disadvantages of the amateur. Specifically, \citet{chuang2024dola} employ the predictive distribution from the higher layer as the expert and that from the lower layer as the amateur, aiming to emphasize the factual knowledge embedded in the higher layer. Similarly, \citet{zhang2023alleviating} utilize the original LLM as the expert and a hallucination-prone LLM as the amateur to amplify the knowledge within the original model and reduce its tendency to fabricate information. In addition to focusing on parametric knowledge, \citet{shi2024trusting} propose context-aware decoding, where both the expert and the amateur share the same LLM, but only the expert has access to context, thereby highlighting the importance of contextual knowledge in responses. The idea of contrastive decoding is also applicable in multimodal scenarios. \citet{leng2024mitigating} use a model with original visual inputs as the expert and the same model with distorted visual inputs as the amateur, with the goal of highlighting the role of visual inputs in shaping responses.


\textit{Summary \& Discussion.} Decoding-time intervention shows great promise in unlocking the potential of LLMs, but current research faces two primary challenges. First, the generalization is limited, as most methods have been concentrated on specific domains, such as factuality or reasoning, leaving their effectiveness in general instruction-following scenarios relatively underexplored. Additionally, these approaches may incur extra computational costs, such as an extra forward pass in contrastive decoding. Future research could develop strategies to address these challenges.

\paragraph{Sampling and Aggregation.}
A straightforward approach for eliciting faithful knowledge from LLMs is through sampling and aggregation. This involves sampling multiple outputs and then aggregating them to derive the most consistent one, which is expected to more accurately reflect the model's knowledge. \citet{wang2023selfconsistency} sample a set of reasoning paths for a single query and then aggregate their answers by majority voting. This simple strategy achieves great performance in various reasoning tasks. Instead of aggregating answers, \citet{chen2023universal} prompt an LLM to aggregate multiple free-form responses based on majority consensus for open-ended generation. To make the most of multiple outputs, \citet{thirukovalluru2024atomic} split each output into several atomic parts, cluster them using sentence embeddings, remove clusters with fewer elements, and summarize the remaining clusters to produce a final consistent output. Similarly, \citet{wang2024integrate} propose prompting an LLM to integrate and derive the final output based on the majority consensus from multiple outputs.

\textit{Summary \& Discussion.} Intuitively, consistent content across multiple generations tends to faithfully reflect the model’s knowledge. While this approach demonstrates strong performance, the multiple sampling process incurs substantial computational costs, limiting its applicability in real-world settings. To overcome this challenge, future research could focus on constructing training data through sampling and aggregation, then fine-tuning the model with this data to internalize this capability.

\paragraph{Post-generation Revision.}
Another approach involves post-generation refinement, where the response is modified to reduce inconsistencies with the model's knowledge. \citet{dhuliawala2023chain} first prompt the LLM to generate a list of questions designed to verify the atomic facts in its initial response, and then provide answers to these questions individually. Following this, the LLM is prompted to check the consistency between the initial response and the answers, making any necessary revisions. Similarly, \citet{varshney2023stitch} identify key concepts from the response, such as entities and keywords, evaluate their confidence using token probability, and retrieve external knowledge to validate and revise low-confidence concepts, which are more likely to be fabricated. Analogously, \citet{zhao2023verify} assess the consistency across multiple responses as described by \citet{wang2023selfconsistency}, then revise the less consistent ones using external knowledge.

\textit{Summary \& Discussion.} Post-generation revision offers an additional chance to correct inaccurate knowledge expression. The primary concern, however, is the increased computational cost. A possible direction for future research is to create training data based on this strategy and then fine-tune the LLM accordingly. For example, the original outputs and their revised versions could be treated as preference pairs, allowing methods like DPO or PPO to align the LLM with the desired attributes of the revised outputs.

\subsection{Training-based Approaches}
\label{sec:improvement_of_imporving_self_consistency_training_based}

\paragraph{Self-aware Fine-tuning.}
Recent research suggests that fine-tuning LLMs with new knowledge may diminish their ability to express knowledge accurately, as this process can teach them to fabricate information beyond their internal knowledge \citep{gudibande2023false, lin2024flame, gekhman2024does}. To address this issue, many studies have started taking the knowledge boundaries of LLMs into account during fine-tuning, an approach we refer to as self-aware fine-tuning. \citet{yang2023alignment, zhang2024r, cheng2024can} fine-tune LLMs to explicitly state \textit{``I don’t know''} when they lack adequate knowledge, thereby reducing the risk of generating fabricated information. Alternatively, \citet{wan2024knowledge} propose discard tuning, where examples are discarded when the model lacks the necessary knowledge, and open-book tuning, which incorporates reference knowledge into the context during fine-tuning to prevent the models from learning to fabricate content. More details on the knowledge identification methods used in these studies can be found in \textsection\ref{sec:improvement_of_self_knowledge_training_based}. \citet{kang2024unfamiliar} introduce a conservative reward model that encourages less detailed responses in situations where the LLM is unfamiliar with the queries.

\textit{Summary \& Discussion.}
Self-aware fine-tuning has demonstrated great potential in alleviating the tendency to fabricate information. The primary challenge lies in distinguishing between what the model knows and doesn’t know. For future research, RL-based self-ware fine-tuning could be a valuable area for exploration, as it allows the LLM to explore broader spaces during training, potentially providing a more accurate reflection of its knowledge boundary.

\paragraph{Self-supervised Fine-tuning.}
Another line of research fine-tunes LLMs to improve the expression of knowledge by leveraging supervision from their internal knowledge. \citet{tian2024finetuning} initiate this process by prompting GPT-3.5 to extract multiple atomic claims from long-form generations. For each claim, they create a verification question, sample multiple answers from the LLM, and then calculate the consistency score across these answers. Based on the average consistency score of all claims, they construct preference pairs and employ DPO for optimization. Similarly, \citet{zhang2024self} propose a method that use self-evaluation to verify each claim and also utilizes DPO for optimization. Simply, \citet{lin2024flame} prompt the LLM to generate responses for fact-based instructions, which are then used to fine-tune the model. After the fine-tuning phase, they create preference pairs by sampling responses from the LLM and use the model itself to assess these responses. Finally, they apply DPO for further refinement. 

\textit{Summary \& Discussion.} Self-supervised fine-tuning has proven effective in improving LLMs' knowledge expression ability. However, a major concern is the quality of self-provided supervision signals, as these models may produce incorrect statements, and recent studies \citep{huang2024large, jiang2024self} have raised doubts about their self-evaluation abilities. Future research could aim to more thoroughly investigate the reliability of self-supervised fine-tuning.

\paragraph{Others.}
In addition to the previously mentioned methods, there are other techniques aimed at improving the ability of LLMs to express their knowledge. \citet{wei2023simple} construct simple synthetic data to train LLMs to avoid sycophancy, they reformat existing publicly available NLP datasets by adding user opinions that are independent of the correctness of the final answer and then fine-tune the models using this data. To achieve consistent model outputs across various prompts, \citet{zhou2022prompt, cao2024worst} apply a consistency loss, which regularizes the outputs of semantically equivalent prompts to remain the same.

\section{Future Work}
\label{sec:future_work}

In this section, we discuss several unresolved research challenges associated with honesty and provide insights into potential research avenues.

\paragraph{Objective or Subjective.} A central debate in current research on the honesty of LLMs revolves around whether honesty should be considered a subjective or objective concept. \citet{askell2021general, kadavath2022language} describe honesty as the ability to provide accurate information along with calibrated confidence reflecting the correctness of its answers. In contrast, \citet{evans2021truthful, lin2022teaching} view honesty as the model's ability to express its own beliefs. The former takes an objective approach, aligning with world knowledge, while the latter adopts a subjective perspective, focusing on the model’s internal state. The objective perspective better suits human needs, as people generally value accurate and truthful information. However, this approach presents challenges for optimizing models because there is often a gap between the model’s knowledge and world knowledge, necessitating additional supervision during training to \textit{distinguish between true and false information}. This becomes particularly challenging for future superhuman models, where human oversight may be limited to providing only weak supervision \citep{burns2023weak}. Conversely, the subjective perspective might require less supervision, as it primarily focuses on the model’s ability to express its own knowledge, but the challenge lies in \textit{differentiating between known and unknown knowledge}. Nonetheless, even if a model can fully articulate its knowledge, problems arise when that knowledge is incorrect. Both the objective and subjective perspectives have distinct advantages and challenges, and resolving this debate is crucial for further research progress.

\paragraph{Knowledge Identification.} As stated in previous sections, knowledge identification has been the primary challenge in both evaluation and methodological approaches. However, the exact definitions of what should be considered known or unknown remain unclear. Existing research typically follows two mainstream strategies: a supervised approach, which distinguishes them based on the correctness of responses, and an unsupervised approach, which differentiates them based on the uncertainty in responses. Both strategies depend on the external expression of LLMs, but what if these models struggle to express what they know? Indeed, several studies have highlighted the discrepancy between LLMs' internal knowledge and what they express \citep{liu2023cognitive, li2024inference}. Ignoring this issue could limit the potential of LLMs to fully express their knowledge. Therefore, in addition to focusing on external expression, future research could explore techniques that utilize the inherent knowledge LLMs possess to differentiate between known and unknown information, such as the knowledge embedded in their parameters or present in the contexts.

\paragraph{Honesty in Instruction-following.}
Current research on honesty focuses primarily on knowledge-intensive question answering, particularly those involving short-form answers, while largely overlooking instruction-following scenarios that are more desired in real-world applications. Instruction-following differs from question answering as it requires broader capabilities and typically involves long-form generation. To address this gap, future research can establish evaluation methods and benchmarks to assess the honesty of LLMs in instruction-following. Additionally, they can explore methods for improvement, such as prompting, supervised fine-tuning, and reinforcement learning. 

\paragraph{Honesty on In-context Knowledge.}
As noted in \textsection{\ref{sec:what_is_honesty_in_llms}}, LLMs possess two types of knowledge: internal parametric knowledge acquired through training and external in-context knowledge. While most existing research emphasizes the honesty of parametric knowledge, the honesty of in-context knowledge has received less attention. However, in real-world applications of LLMs, in-context knowledge also plays a vital role in generation, particularly in retrieval-augmented and long-context scenarios. Therefore, we advocate for future research to devote more attention to the honesty of in-context knowledge.

\paragraph{Honesty in Various Models.}
Research on honesty has primarily focused on transformer decoder-based LLMs. However, other popular models also deserve attention, including multimodal LLMs (e.g., GPT-4V \citep{achiam2023gpt} and Gemini \citep{team2023gemini}), models with novel architectures (e.g., Mamba \citep{gu2023mamba}), and compressed models using compression techniques such as quantization and pruning \citep{wan2024efficient}. We believe that these models also merit further exploration in future research.





\section{Conclusion}
Honesty is a crucial factor in the development of LLMs, yet current models still exhibit significant dishonest behaviors. To address these issues, this paper offers a thorough overview of research on the honesty of LLMs, including its clarification, evaluation approaches, and improvement strategies. Furthermore, we propose several potential directions for future research. We hope this survey serves as a valuable resource for researchers studying LLM honesty and encourages further exploration in this field.

\bibliography{main}

\begin{thebibliography}{155}
\providecommand{\natexlab}[1]{#1}
\providecommand{\url}[1]{\texttt{#1}}
\expandafter\ifx\csname urlstyle\endcsname\relax
  \providecommand{\doi}[1]{doi: #1}\else
  \providecommand{\doi}{doi: \begingroup \urlstyle{rm}\Url}\fi

\bibitem[Achiam et~al.(2023)Achiam, Adler, Agarwal, Ahmad, Akkaya, Aleman, Almeida, Altenschmidt, Altman, Anadkat, et~al.]{achiam2023gpt}
Josh Achiam, Steven Adler, Sandhini Agarwal, Lama Ahmad, Ilge Akkaya, Florencia~Leoni Aleman, Diogo Almeida, Janko Altenschmidt, Sam Altman, Shyamal Anadkat, et~al.
\newblock Gpt-4 technical report.
\newblock \emph{arXiv preprint arXiv:2303.08774}, 2023.

\bibitem[Adiwardana et~al.(2020)Adiwardana, Luong, So, Hall, Fiedel, Thoppilan, Yang, Kulshreshtha, Nemade, Lu, et~al.]{adiwardana2020towards}
Daniel Adiwardana, Minh-Thang Luong, David~R So, Jamie Hall, Noah Fiedel, Romal Thoppilan, Zi~Yang, Apoorv Kulshreshtha, Gaurav Nemade, Yifeng Lu, et~al.
\newblock Towards a human-like open-domain chatbot.
\newblock \emph{arXiv preprint arXiv:2001.09977}, 2020.

\bibitem[Alain \& Bengio(2016)Alain and Bengio]{alain2016understanding}
Guillaume Alain and Yoshua Bengio.
\newblock Understanding intermediate layers using linear classifier probes.
\newblock \emph{arXiv preprint arXiv:1610.01644}, 2016.

\bibitem[Amayuelas et~al.(2023)Amayuelas, Pan, Chen, and Wang]{amayuelas2023knowledge}
Alfonso Amayuelas, Liangming Pan, Wenhu Chen, and William Wang.
\newblock Knowledge of knowledge: Exploring known-unknowns uncertainty with large language models.
\newblock \emph{arXiv preprint arXiv:2305.13712}, 2023.

\bibitem[Angelopoulos et~al.(2024)Angelopoulos, Bates, Fisch, Lei, and Schuster]{angelopoulos2024conformal}
Anastasios~Nikolas Angelopoulos, Stephen Bates, Adam Fisch, Lihua Lei, and Tal Schuster.
\newblock Conformal risk control.
\newblock In \emph{The Twelfth International Conference on Learning Representations}, 2024.

\bibitem[Askell et~al.(2021)Askell, Bai, Chen, Drain, Ganguli, Henighan, Jones, Joseph, Mann, DasSarma, et~al.]{askell2021general}
Amanda Askell, Yuntao Bai, Anna Chen, Dawn Drain, Deep Ganguli, Tom Henighan, Andy Jones, Nicholas Joseph, Ben Mann, Nova DasSarma, et~al.
\newblock A general language assistant as a laboratory for alignment.
\newblock \emph{arXiv preprint arXiv:2112.00861}, 2021.

\bibitem[Azaria \& Mitchell(2023)Azaria and Mitchell]{azaria2023internal}
Amos Azaria and Tom Mitchell.
\newblock The internal state of an llm knows when it’s lying.
\newblock In \emph{Findings of the Association for Computational Linguistics: EMNLP 2023}, pp.\  967--976, 2023.

\bibitem[Bai et~al.(2022)Bai, Jones, Ndousse, Askell, Chen, DasSarma, Drain, Fort, Ganguli, Henighan, et~al.]{bai2022training}
Yuntao Bai, Andy Jones, Kamal Ndousse, Amanda Askell, Anna Chen, Nova DasSarma, Dawn Drain, Stanislav Fort, Deep Ganguli, Tom Henighan, et~al.
\newblock Training a helpful and harmless assistant with reinforcement learning from human feedback.
\newblock \emph{arXiv preprint arXiv:2204.05862}, 2022.

\bibitem[Bai et~al.(2024)Bai, Wang, Xiao, He, Han, Zhang, and Shou]{bai2024hallucination}
Zechen Bai, Pichao Wang, Tianjun Xiao, Tong He, Zongbo Han, Zheng Zhang, and Mike~Zheng Shou.
\newblock Hallucination of multimodal large language models: A survey.
\newblock \emph{arXiv preprint arXiv:2404.18930}, 2024.

\bibitem[Band et~al.(2024)Band, Li, Ma, and Hashimoto]{band2024linguistic}
Neil Band, Xuechen Li, Tengyu Ma, and Tatsunori Hashimoto.
\newblock Linguistic calibration of long-form generations.
\newblock In \emph{Forty-first International Conference on Machine Learning}, 2024.

\bibitem[Belinkov(2022)]{belinkov2022probing}
Yonatan Belinkov.
\newblock Probing classifiers: Promises, shortcomings, and advances.
\newblock \emph{Computational Linguistics}, 48\penalty0 (1):\penalty0 207--219, 2022.

\bibitem[Brier(1950)]{brier1950verification}
Glenn~W Brier.
\newblock Verification of forecasts expressed in terms of probability.
\newblock \emph{Monthly weather review}, 78\penalty0 (1):\penalty0 1--3, 1950.

\bibitem[Burns et~al.(2023{\natexlab{a}})Burns, Izmailov, Kirchner, Baker, Gao, Aschenbrenner, Chen, Ecoffet, Joglekar, Leike, et~al.]{burns2023weak}
Collin Burns, Pavel Izmailov, Jan~Hendrik Kirchner, Bowen Baker, Leo Gao, Leopold Aschenbrenner, Yining Chen, Adrien Ecoffet, Manas Joglekar, Jan Leike, et~al.
\newblock Weak-to-strong generalization: Eliciting strong capabilities with weak supervision.
\newblock \emph{arXiv preprint arXiv:2312.09390}, 2023{\natexlab{a}}.

\bibitem[Burns et~al.(2023{\natexlab{b}})Burns, Ye, Klein, and Steinhardt]{burns2023discovering}
Collin Burns, Haotian Ye, Dan Klein, and Jacob Steinhardt.
\newblock Discovering latent knowledge in language models without supervision.
\newblock In \emph{The Eleventh International Conference on Learning Representations}, 2023{\natexlab{b}}.

\bibitem[Cao et~al.(2024)Cao, Cai, Zhang, Zou, and Lam]{cao2024worst}
Bowen Cao, Deng Cai, Zhisong Zhang, Yuexian Zou, and Wai Lam.
\newblock On the worst prompt performance of large language models.
\newblock \emph{arXiv preprint arXiv:2406.10248}, 2024.

\bibitem[Carlini et~al.(2023)Carlini, Ippolito, Jagielski, Lee, Tramer, and Zhang]{carlini2023quantifying}
Nicholas Carlini, Daphne Ippolito, Matthew Jagielski, Katherine Lee, Florian Tramer, and Chiyuan Zhang.
\newblock Quantifying memorization across neural language models.
\newblock In \emph{The Eleventh International Conference on Learning Representations}, 2023.

\bibitem[Chen et~al.(2024{\natexlab{a}})Chen, Liu, Chen, Gu, Wu, Tao, Fu, and Ye]{chen2024inside}
Chao Chen, Kai Liu, Ze~Chen, Yi~Gu, Yue Wu, Mingyuan Tao, Zhihang Fu, and Jieping Ye.
\newblock {INSIDE}: {LLM}s' internal states retain the power of hallucination detection.
\newblock In \emph{The Twelfth International Conference on Learning Representations}, 2024{\natexlab{a}}.

\bibitem[Chen et~al.(2023{\natexlab{a}})Chen, Yoon, Ebrahimi, Arik, Pfister, and Jha]{chen2023adaptation}
Jiefeng Chen, Jinsung Yoon, Sayna Ebrahimi, Sercan Arik, Tomas Pfister, and Somesh Jha.
\newblock Adaptation with self-evaluation to improve selective prediction in llms.
\newblock In \emph{Findings of the Association for Computational Linguistics: EMNLP 2023}, pp.\  5190--5213, 2023{\natexlab{a}}.

\bibitem[Chen et~al.(2024{\natexlab{b}})Chen, Liang, Wang, Liang, Xiao, Wei, Chen, Hao, Han, and Wang]{chen2024teaching}
Lida Chen, Zujie Liang, Xintao Wang, Jiaqing Liang, Yanghua Xiao, Feng Wei, Jinglei Chen, Zhenghong Hao, Bing Han, and Wei Wang.
\newblock Teaching large language models to express knowledge boundary from their own signals.
\newblock \emph{arXiv preprint arXiv:2406.10881}, 2024{\natexlab{b}}.

\bibitem[Chen et~al.(2024{\natexlab{c}})Chen, Xiong, Liu, Wu, Xiao, Gao, and He]{chen2024context}
Shiqi Chen, Miao Xiong, Junteng Liu, Zhengxuan Wu, Teng Xiao, Siyang Gao, and Junxian He.
\newblock In-context sharpness as alerts: An inner representation perspective for hallucination mitigation.
\newblock In \emph{Forty-first International Conference on Machine Learning}, 2024{\natexlab{c}}.

\bibitem[Chen et~al.(2023{\natexlab{b}})Chen, Aksitov, Alon, Ren, Xiao, Yin, Prakash, Sutton, Wang, and Zhou]{chen2023universal}
Xinyun Chen, Renat Aksitov, Uri Alon, Jie Ren, Kefan Xiao, Pengcheng Yin, Sushant Prakash, Charles Sutton, Xuezhi Wang, and Denny Zhou.
\newblock Universal self-consistency for large language model generation.
\newblock \emph{arXiv preprint arXiv:2311.17311}, 2023{\natexlab{b}}.

\bibitem[Cheng et~al.(2024)Cheng, Sun, Liu, Zhang, Yin, Li, Li, Chen, and Qiu]{cheng2024can}
Qinyuan Cheng, Tianxiang Sun, Xiangyang Liu, Wenwei Zhang, Zhangyue Yin, Shimin Li, Linyang Li, Kai Chen, and Xipeng Qiu.
\newblock Can ai assistants know what they don't know?
\newblock In \emph{Forty-first International Conference on Machine Learning}, 2024.

\bibitem[Chern et~al.(2024)Chern, Hu, Yang, Chern, Guo, Jin, Wang, and Liu]{chern2024behonest}
Steffi Chern, Zhulin Hu, Yuqing Yang, Ethan Chern, Yuan Guo, Jiahe Jin, Binjie Wang, and Pengfei Liu.
\newblock Behonest: Benchmarking honesty of large language models.
\newblock \emph{arXiv preprint arXiv:2406.13261}, 2024.

\bibitem[Chuang et~al.(2024)Chuang, Xie, Luo, Kim, Glass, and He]{chuang2024dola}
Yung-Sung Chuang, Yujia Xie, Hongyin Luo, Yoon Kim, James~R. Glass, and Pengcheng He.
\newblock Dola: Decoding by contrasting layers improves factuality in large language models.
\newblock In \emph{The Twelfth International Conference on Learning Representations}, 2024.

\bibitem[Cobbe et~al.(2021)Cobbe, Kosaraju, Bavarian, Chen, Jun, Kaiser, Plappert, Tworek, Hilton, Nakano, et~al.]{cobbe2021training}
Karl Cobbe, Vineet Kosaraju, Mohammad Bavarian, Mark Chen, Heewoo Jun, Lukasz Kaiser, Matthias Plappert, Jerry Tworek, Jacob Hilton, Reiichiro Nakano, et~al.
\newblock Training verifiers to solve math word problems.
\newblock \emph{arXiv preprint arXiv:2110.14168}, 2021.

\bibitem[Dahl et~al.(2024)Dahl, Magesh, Suzgun, and Ho]{dahl2024large}
Matthew Dahl, Varun Magesh, Mirac Suzgun, and Daniel~E Ho.
\newblock Large legal fictions: Profiling legal hallucinations in large language models.
\newblock \emph{arXiv preprint arXiv:2401.01301}, 2024.

\bibitem[DeGroot \& Fienberg(1983)DeGroot and Fienberg]{degroot1983comparison}
Morris~H DeGroot and Stephen~E Fienberg.
\newblock The comparison and evaluation of forecasters.
\newblock \emph{Journal of the Royal Statistical Society: Series D (The Statistician)}, 32\penalty0 (1-2):\penalty0 12--22, 1983.

\bibitem[Dhuliawala et~al.(2023)Dhuliawala, Komeili, Xu, Raileanu, Li, Celikyilmaz, and Weston]{dhuliawala2023chain}
Shehzaad Dhuliawala, Mojtaba Komeili, Jing Xu, Roberta Raileanu, Xian Li, Asli Celikyilmaz, and Jason Weston.
\newblock Chain-of-verification reduces hallucination in large language models.
\newblock \emph{arXiv preprint arXiv:2309.11495}, 2023.

\bibitem[Dictionary(1989)]{dictionary1989oxford}
Oxford~English Dictionary.
\newblock Oxford english dictionary.
\newblock \emph{Simpson, Ja \& Weiner, Esc}, 3, 1989.

\bibitem[Duan et~al.(2024)Duan, Cheng, Wang, Zavalny, Wang, Xu, Kailkhura, and Xu]{duan2024shifting}
Jinhao Duan, Hao Cheng, Shiqi Wang, Alex Zavalny, Chenan Wang, Renjing Xu, Bhavya Kailkhura, and Kaidi Xu.
\newblock Shifting attention to relevance: Towards the predictive uncertainty quantification of free-form large language models.
\newblock In Lun-Wei Ku, Andre Martins, and Vivek Srikumar (eds.), \emph{Proceedings of the 62nd Annual Meeting of the Association for Computational Linguistics (Volume 1: Long Papers)}, pp.\  5050--5063, 2024.

\bibitem[Evans et~al.(2021)Evans, Cotton-Barratt, Finnveden, Bales, Balwit, Wills, Righetti, and Saunders]{evans2021truthful}
Owain Evans, Owen Cotton-Barratt, Lukas Finnveden, Adam Bales, Avital Balwit, Peter Wills, Luca Righetti, and William Saunders.
\newblock Truthful ai: Developing and governing ai that does not lie.
\newblock \emph{arXiv preprint arXiv:2110.06674}, 2021.

\bibitem[Fadeeva et~al.(2024)Fadeeva, Rubashevskii, Shelmanov, Petrakov, Li, Mubarak, Tsymbalov, Kuzmin, Panchenko, Baldwin, et~al.]{fadeeva2024fact}
Ekaterina Fadeeva, Aleksandr Rubashevskii, Artem Shelmanov, Sergey Petrakov, Haonan Li, Hamdy Mubarak, Evgenii Tsymbalov, Gleb Kuzmin, Alexander Panchenko, Timothy Baldwin, et~al.
\newblock Fact-checking the output of large language models via token-level uncertainty quantification.
\newblock In \emph{Findings of the Association for Computational Linguistics ACL 2024}, pp.\  9367--9385, 2024.

\bibitem[Farquhar et~al.(2023)Farquhar, Varma, Kenton, Gasteiger, Mikulik, and Shah]{farquhar2023challenges}
Sebastian Farquhar, Vikrant Varma, Zachary Kenton, Johannes Gasteiger, Vladimir Mikulik, and Rohin Shah.
\newblock Challenges with unsupervised llm knowledge discovery.
\newblock \emph{arXiv preprint arXiv:2312.10029}, 2023.

\bibitem[Farquhar et~al.(2024)Farquhar, Kossen, Kuhn, and Gal]{farquhar2024detecting}
Sebastian Farquhar, Jannik Kossen, Lorenz Kuhn, and Yarin Gal.
\newblock Detecting hallucinations in large language models using semantic entropy.
\newblock \emph{Nature}, 630\penalty0 (8017):\penalty0 625--630, 2024.

\bibitem[Feng et~al.(2024)Feng, Shi, Wang, Ding, Balachandran, and Tsvetkov]{feng-etal-2024-dont}
Shangbin Feng, Weijia Shi, Yike Wang, Wenxuan Ding, Vidhisha Balachandran, and Yulia Tsvetkov.
\newblock Don{'}t hallucinate, abstain: Identifying {LLM} knowledge gaps via multi-{LLM} collaboration.
\newblock In \emph{Proceedings of the 62nd Annual Meeting of the Association for Computational Linguistics (Volume 1: Long Papers)}, pp.\  14664--14690, 2024.

\bibitem[Gao et~al.(2024)Gao, Zhang, Chen, Huang, Wu, Fu, Wan, Zhang, and Sun]{gao2024best}
Chujie Gao, Qihui Zhang, Dongping Chen, Yue Huang, Siyuan Wu, Zhengyan Fu, Yao Wan, Xiangliang Zhang, and Lichao Sun.
\newblock The best of both worlds: Toward an honest and helpful large language model.
\newblock \emph{arXiv preprint arXiv:2406.00380}, 2024.

\bibitem[Gekhman et~al.(2024)Gekhman, Yona, Aharoni, Eyal, Feder, Reichart, and Herzig]{gekhman2024does}
Zorik Gekhman, Gal Yona, Roee Aharoni, Matan Eyal, Amir Feder, Roi Reichart, and Jonathan Herzig.
\newblock Does fine-tuning llms on new knowledge encourage hallucinations?
\newblock \emph{arXiv preprint arXiv:2405.05904}, 2024.

\bibitem[Geng et~al.(2024)Geng, Cai, Wang, Koeppl, Nakov, and Gurevych]{geng2024survey}
Jiahui Geng, Fengyu Cai, Yuxia Wang, Heinz Koeppl, Preslav Nakov, and Iryna Gurevych.
\newblock A survey of confidence estimation and calibration in large language models.
\newblock In \emph{Proceedings of the 2024 Conference of the North American Chapter of the Association for Computational Linguistics: Human Language Technologies (Volume 1: Long Papers)}, pp.\  6577--6595, 2024.

\bibitem[Gertler(2010)]{gertler2010self}
Brie Gertler.
\newblock \emph{Self-knowledge}.
\newblock Routledge, 2010.

\bibitem[Gu \& Dao(2023)Gu and Dao]{gu2023mamba}
Albert Gu and Tri Dao.
\newblock Mamba: Linear-time sequence modeling with selective state spaces.
\newblock \emph{arXiv preprint arXiv:2312.00752}, 2023.

\bibitem[Gu et~al.(2023)Gu, Zhao, Xu, Nie, Mei, and Yin]{gu2022robustness}
Jiasheng Gu, Hongyu Zhao, Hanzi Xu, Liangyu Nie, Hongyuan Mei, and Wenpeng Yin.
\newblock Robustness of learning from task instructions.
\newblock In \emph{Findings of the Association for Computational Linguistics: ACL 2023}, 2023.

\bibitem[Gudibande et~al.(2023)Gudibande, Wallace, Snell, Geng, Liu, Abbeel, Levine, and Song]{gudibande2023false}
Arnav Gudibande, Eric Wallace, Charlie Snell, Xinyang Geng, Hao Liu, Pieter Abbeel, Sergey Levine, and Dawn Song.
\newblock The false promise of imitating proprietary llms.
\newblock \emph{arXiv preprint arXiv:2305.15717}, 2023.

\bibitem[Guo et~al.(2017)Guo, Pleiss, Sun, and Weinberger]{guo2017calibration}
Chuan Guo, Geoff Pleiss, Yu~Sun, and Kilian~Q Weinberger.
\newblock On calibration of modern neural networks.
\newblock In \emph{International conference on machine learning}, pp.\  1321--1330. PMLR, 2017.

\bibitem[Gupta et~al.(2024)Gupta, Narasimhan, Jitkrittum, Rawat, Menon, and Kumar]{gupta2024language}
Neha Gupta, Harikrishna Narasimhan, Wittawat Jitkrittum, Ankit~Singh Rawat, Aditya~Krishna Menon, and Sanjiv Kumar.
\newblock Language model cascades: Token-level uncertainty and beyond.
\newblock \emph{arXiv preprint arXiv:2404.10136}, 2024.

\bibitem[Han et~al.(2024)Han, Li, Chen, Shi, Du, Xiao, Liang, and Lin]{han2024enhancing}
Haixia Han, Tingyun Li, Shisong Chen, Jie Shi, Chengyu Du, Yanghua Xiao, Jiaqing Liang, and Xin Lin.
\newblock Enhancing confidence expression in large language models through learning from past experience.
\newblock \emph{arXiv preprint arXiv:2404.10315}, 2024.

\bibitem[Hu et~al.(2021)Hu, Shen, Wallis, Allen-Zhu, Li, Wang, Wang, and Chen]{hu2021lora}
Edward~J Hu, Yelong Shen, Phillip Wallis, Zeyuan Allen-Zhu, Yuanzhi Li, Shean Wang, Lu~Wang, and Weizhu Chen.
\newblock Lora: Low-rank adaptation of large language models.
\newblock \emph{arXiv preprint arXiv:2106.09685}, 2021.

\bibitem[Hu et~al.(2023)Hu, Luo, Wang, Cheng, Liu, and Sun]{hu2023won}
Shengding Hu, Yifan Luo, Huadong Wang, Xingyi Cheng, Zhiyuan Liu, and Maosong Sun.
\newblock Won’t get fooled again: Answering questions with false premises.
\newblock In \emph{Proceedings of the 61st Annual Meeting of the Association for Computational Linguistics (Volume 1: Long Papers)}, pp.\  5626--5643, 2023.

\bibitem[Huang et~al.(2024{\natexlab{a}})Huang, Chen, Mishra, Zheng, Yu, Song, and Zhou]{huang2024large}
Jie Huang, Xinyun Chen, Swaroop Mishra, Huaixiu~Steven Zheng, Adams~Wei Yu, Xinying Song, and Denny Zhou.
\newblock Large language models cannot self-correct reasoning yet.
\newblock In \emph{The Twelfth International Conference on Learning Representations}, 2024{\natexlab{a}}.

\bibitem[Huang et~al.(2024{\natexlab{b}})Huang, Sun, Wang, Wu, Zhang, Li, Gao, Huang, Lyu, Zhang, Li, Sun, Liu, Liu, Wang, Zhang, Vidgen, Kailkhura, Xiong, Xiao, Li, Xing, Huang, Liu, Ji, Wang, Zhang, Yao, Kellis, Zitnik, Jiang, Bansal, Zou, Pei, Liu, Gao, Han, Zhao, Tang, Wang, Vanschoren, Mitchell, Shu, Xu, Chang, He, Huang, Backes, Gong, Yu, Chen, Gu, Xu, Ying, Ji, Jana, Chen, Liu, Zhou, Wang, Li, Zhang, Wang, Xie, Chen, Wang, Liu, Ye, Cao, Chen, and Zhao]{huang2024trustllm}
Yue Huang, Lichao Sun, Haoran Wang, Siyuan Wu, Qihui Zhang, Yuan Li, Chujie Gao, Yixin Huang, Wenhan Lyu, Yixuan Zhang, Xiner Li, Hanchi Sun, Zhengliang Liu, Yixin Liu, Yijue Wang, Zhikun Zhang, Bertie Vidgen, Bhavya Kailkhura, Caiming Xiong, Chaowei Xiao, Chunyuan Li, Eric~P. Xing, Furong Huang, Hao Liu, Heng Ji, Hongyi Wang, Huan Zhang, Huaxiu Yao, Manolis Kellis, Marinka Zitnik, Meng Jiang, Mohit Bansal, James Zou, Jian Pei, Jian Liu, Jianfeng Gao, Jiawei Han, Jieyu Zhao, Jiliang Tang, Jindong Wang, Joaquin Vanschoren, John Mitchell, Kai Shu, Kaidi Xu, Kai-Wei Chang, Lifang He, Lifu Huang, Michael Backes, Neil~Zhenqiang Gong, Philip~S. Yu, Pin-Yu Chen, Quanquan Gu, Ran Xu, Rex Ying, Shuiwang Ji, Suman Jana, Tianlong Chen, Tianming Liu, Tianyi Zhou, William~Yang Wang, Xiang Li, Xiangliang Zhang, Xiao Wang, Xing Xie, Xun Chen, Xuyu Wang, Yan Liu, Yanfang Ye, Yinzhi Cao, Yong Chen, and Yue Zhao.
\newblock Trustllm: Trustworthiness in large language models.
\newblock In \emph{Forty-first International Conference on Machine Learning}, 2024{\natexlab{b}}.

\bibitem[Huang et~al.(2024{\natexlab{c}})Huang, Liu, Thirukovalluru, Cohan, and Dhingra]{huang2024calibrating}
Yukun Huang, Yixin Liu, Raghuveer Thirukovalluru, Arman Cohan, and Bhuwan Dhingra.
\newblock Calibrating long-form generations from large language models.
\newblock \emph{arXiv preprint arXiv:2402.06544}, 2024{\natexlab{c}}.

\bibitem[Ji et~al.(2024)Ji, Chen, Ishii, Cahyawijaya, Bang, Wilie, and Fung]{ji2024llm}
Ziwei Ji, Delong Chen, Etsuko Ishii, Samuel Cahyawijaya, Yejin Bang, Bryan Wilie, and Pascale Fung.
\newblock Llm internal states reveal hallucination risk faced with a query.
\newblock \emph{arXiv preprint arXiv:2407.03282}, 2024.

\bibitem[Jiang et~al.(2024)Jiang, Zhang, Weller, Weir, Van~Durme, and Khashabi]{jiang2024self}
Dongwei Jiang, Jingyu Zhang, Orion Weller, Nathaniel Weir, Benjamin Van~Durme, and Daniel Khashabi.
\newblock Self-[in] correct: Llms struggle with refining self-generated responses.
\newblock \emph{arXiv preprint arXiv:2404.04298}, 2024.

\bibitem[Jiang et~al.(2023)Jiang, Xu, Gao, Sun, Liu, Dwivedi-Yu, Yang, Callan, and Neubig]{jiang2023active}
Zhengbao Jiang, Frank~F Xu, Luyu Gao, Zhiqing Sun, Qian Liu, Jane Dwivedi-Yu, Yiming Yang, Jamie Callan, and Graham Neubig.
\newblock Active retrieval augmented generation.
\newblock In \emph{Proceedings of the 2023 Conference on Empirical Methods in Natural Language Processing}, pp.\  7969--7992, 2023.

\bibitem[Joshi et~al.(2017)Joshi, Choi, Weld, and Zettlemoyer]{joshi2017triviaqa}
Mandar Joshi, Eunsol Choi, Daniel~S Weld, and Luke Zettlemoyer.
\newblock Triviaqa: A large scale distantly supervised challenge dataset for reading comprehension.
\newblock In \emph{Proceedings of the 55th Annual Meeting of the Association for Computational Linguistics (Volume 1: Long Papers)}, pp.\  1601--1611, 2017.

\bibitem[Jung et~al.(2024)Jung, Brahman, and Choi]{jung2024trust}
Jaehun Jung, Faeze Brahman, and Yejin Choi.
\newblock Trust or escalate: Llm judges with provable guarantees for human agreement.
\newblock \emph{arXiv preprint arXiv:2407.18370}, 2024.

\bibitem[Kadavath et~al.(2022)Kadavath, Conerly, Askell, Henighan, Drain, Perez, Schiefer, Hatfield-Dodds, DasSarma, Tran-Johnson, et~al.]{kadavath2022language}
Saurav Kadavath, Tom Conerly, Amanda Askell, Tom Henighan, Dawn Drain, Ethan Perez, Nicholas Schiefer, Zac Hatfield-Dodds, Nova DasSarma, Eli Tran-Johnson, et~al.
\newblock Language models (mostly) know what they know.
\newblock \emph{arXiv preprint arXiv:2207.05221}, 2022.

\bibitem[Kang et~al.(2024)Kang, Wallace, Tomlin, Kumar, and Levine]{kang2024unfamiliar}
Katie Kang, Eric Wallace, Claire Tomlin, Aviral Kumar, and Sergey Levine.
\newblock Unfamiliar finetuning examples control how language models hallucinate.
\newblock \emph{arXiv preprint arXiv:2403.05612}, 2024.

\bibitem[Kapoor et~al.(2024)Kapoor, Gruver, Roberts, Collins, Pal, Bhatt, Weller, Dooley, Goldblum, and Wilson]{kapoor2024large}
Sanyam Kapoor, Nate Gruver, Manley Roberts, Katherine Collins, Arka Pal, Umang Bhatt, Adrian Weller, Samuel Dooley, Micah Goldblum, and Andrew~Gordon Wilson.
\newblock Large language models must be taught to know what they don't know.
\newblock \emph{arXiv preprint arXiv:2406.08391}, 2024.

\bibitem[Kim \& Ko(2011)Kim and Ko]{kim2011culture}
Heejung~S Kim and Deborah Ko.
\newblock Culture and self-expression.
\newblock In \emph{The self}, pp.\  325--342. Psychology Press, 2011.

\bibitem[Kojima et~al.(2022)Kojima, Gu, Reid, Matsuo, and Iwasawa]{kojima2022large}
Takeshi Kojima, Shixiang~Shane Gu, Machel Reid, Yutaka Matsuo, and Yusuke Iwasawa.
\newblock Large language models are zero-shot reasoners.
\newblock \emph{Advances in neural information processing systems}, 35:\penalty0 22199--22213, 2022.

\bibitem[Kossen et~al.(2024)Kossen, Han, Razzak, Schut, Malik, and Gal]{kossen2024semantic}
Jannik Kossen, Jiatong Han, Muhammed Razzak, Lisa Schut, Shreshth Malik, and Yarin Gal.
\newblock Semantic entropy probes: Robust and cheap hallucination detection in llms.
\newblock \emph{arXiv preprint arXiv:2406.15927}, 2024.

\bibitem[Kuhn et~al.(2023)Kuhn, Gal, and Farquhar]{kuhn2023semantic}
Lorenz Kuhn, Yarin Gal, and Sebastian Farquhar.
\newblock Semantic uncertainty: Linguistic invariances for uncertainty estimation in natural language generation.
\newblock In \emph{The Eleventh International Conference on Learning Representations}, 2023.

\bibitem[Kull et~al.(2019)Kull, Perello~Nieto, K{\"a}ngsepp, Silva~Filho, Song, and Flach]{kull2019beyond}
Meelis Kull, Miquel Perello~Nieto, Markus K{\"a}ngsepp, Telmo Silva~Filho, Hao Song, and Peter Flach.
\newblock Beyond temperature scaling: Obtaining well-calibrated multi-class probabilities with dirichlet calibration.
\newblock \emph{Advances in neural information processing systems}, 32, 2019.

\bibitem[Kwiatkowski et~al.(2019)Kwiatkowski, Palomaki, Redfield, Collins, Parikh, Alberti, Epstein, Polosukhin, Devlin, Lee, et~al.]{kwiatkowski2019natural}
Tom Kwiatkowski, Jennimaria Palomaki, Olivia Redfield, Michael Collins, Ankur Parikh, Chris Alberti, Danielle Epstein, Illia Polosukhin, Jacob Devlin, Kenton Lee, et~al.
\newblock Natural questions: a benchmark for question answering research.
\newblock \emph{Transactions of the Association for Computational Linguistics}, 7:\penalty0 453--466, 2019.

\bibitem[Lachmy et~al.(2022)Lachmy, Pyatkin, Manevich, and Tsarfaty]{lachmy2022draw}
Royi Lachmy, Valentina Pyatkin, Avshalom Manevich, and Reut Tsarfaty.
\newblock Draw me a flower: Processing and grounding abstraction in natural language.
\newblock \emph{Transactions of the Association for Computational Linguistics}, 10:\penalty0 1341--1356, 2022.

\bibitem[Leng et~al.(2024)Leng, Zhang, Chen, Li, Lu, Miao, and Bing]{leng2024mitigating}
Sicong Leng, Hang Zhang, Guanzheng Chen, Xin Li, Shijian Lu, Chunyan Miao, and Lidong Bing.
\newblock Mitigating object hallucinations in large vision-language models through visual contrastive decoding.
\newblock In \emph{Proceedings of the IEEE/CVF Conference on Computer Vision and Pattern Recognition}, pp.\  13872--13882, 2024.

\bibitem[Levinstein \& Herrmann(2024)Levinstein and Herrmann]{levinstein2024still}
Benjamin~A Levinstein and Daniel~A Herrmann.
\newblock Still no lie detector for language models: Probing empirical and conceptual roadblocks.
\newblock \emph{Philosophical Studies}, pp.\  1--27, 2024.

\bibitem[Li et~al.(2024{\natexlab{a}})Li, Patel, Vi{\'e}gas, Pfister, and Wattenberg]{li2024inference}
Kenneth Li, Oam Patel, Fernanda Vi{\'e}gas, Hanspeter Pfister, and Martin Wattenberg.
\newblock Inference-time intervention: Eliciting truthful answers from a language model.
\newblock \emph{Advances in Neural Information Processing Systems}, 36, 2024{\natexlab{a}}.

\bibitem[Li et~al.(2023{\natexlab{a}})Li, Holtzman, Fried, Liang, Eisner, Hashimoto, Zettlemoyer, and Lewis]{li2023contrastive}
Xiang~Lisa Li, Ari Holtzman, Daniel Fried, Percy Liang, Jason Eisner, Tatsunori~B Hashimoto, Luke Zettlemoyer, and Mike Lewis.
\newblock Contrastive decoding: Open-ended text generation as optimization.
\newblock In \emph{Proceedings of the 61st Annual Meeting of the Association for Computational Linguistics (Volume 1: Long Papers)}, pp.\  12286--12312, 2023{\natexlab{a}}.

\bibitem[Li et~al.(2024{\natexlab{b}})Li, Shrivastava, Li, Hashimoto, and Liang]{li2024benchmarking}
Xiang~Lisa Li, Vaishnavi Shrivastava, Siyan Li, Tatsunori Hashimoto, and Percy Liang.
\newblock Benchmarking and improving generator-validator consistency of language models.
\newblock In \emph{The Twelfth International Conference on Learning Representations}, 2024{\natexlab{b}}.

\bibitem[Li et~al.(2023{\natexlab{b}})Li, Wang, Ding, and Chen]{li2023large}
Yinheng Li, Shaofei Wang, Han Ding, and Hang Chen.
\newblock Large language models in finance: A survey.
\newblock In \emph{Proceedings of the fourth ACM international conference on AI in finance}, pp.\  374--382, 2023{\natexlab{b}}.

\bibitem[Lin et~al.(2024{\natexlab{a}})Lin, Gao, Oguz, Xiong, Lin, Yih, and Chen]{lin2024flame}
Sheng-Chieh Lin, Luyu Gao, Barlas Oguz, Wenhan Xiong, Jimmy Lin, Wen-tau Yih, and Xilun Chen.
\newblock Flame: Factuality-aware alignment for large language models.
\newblock \emph{arXiv preprint arXiv:2405.01525}, 2024{\natexlab{a}}.

\bibitem[Lin et~al.(2022{\natexlab{a}})Lin, Hilton, and Evans]{lin2022teaching}
Stephanie Lin, Jacob Hilton, and Owain Evans.
\newblock Teaching models to express their uncertainty in words.
\newblock \emph{Transactions on Machine Learning Research}, 2022{\natexlab{a}}.
\newblock ISSN 2835-8856.

\bibitem[Lin et~al.(2022{\natexlab{b}})Lin, Hilton, and Evans]{lin2022truthfulqa}
Stephanie Lin, Jacob Hilton, and Owain Evans.
\newblock Truthfulqa: Measuring how models mimic human falsehoods.
\newblock In \emph{Proceedings of the 60th Annual Meeting of the Association for Computational Linguistics (Volume 1: Long Papers)}, pp.\  3214--3252, 2022{\natexlab{b}}.

\bibitem[Lin et~al.(2024{\natexlab{b}})Lin, Trivedi, and Sun]{lin2024generating}
Zhen Lin, Shubhendu Trivedi, and Jimeng Sun.
\newblock Generating with confidence: Uncertainty quantification for black-box large language models.
\newblock \emph{Transactions on Machine Learning Research}, 2024{\natexlab{b}}.
\newblock ISSN 2835-8856.

\bibitem[Liu et~al.(2024{\natexlab{a}})Liu, Wang, Yuan, Chen, and Peng]{liu2024examining}
Genglin Liu, Xingyao Wang, Lifan Yuan, Yangyi Chen, and Hao Peng.
\newblock Examining llms' uncertainty expression towards questions outside parametric knowledge, 2024{\natexlab{a}}.

\bibitem[Liu et~al.(2024{\natexlab{b}})Liu, Zhang, Guo, Dong, Li, Lee, Zhang, and Liu]{liu2024ctrla}
Huanshuo Liu, Hao Zhang, Zhijiang Guo, Kuicai Dong, Xiangyang Li, Yi~Quan Lee, Cong Zhang, and Yong Liu.
\newblock Ctrla: Adaptive retrieval-augmented generation via probe-guided control.
\newblock \emph{arXiv preprint arXiv:2405.18727}, 2024{\natexlab{b}}.

\bibitem[Liu et~al.(2024{\natexlab{c}})Liu, Chen, Cheng, and He]{liu2024universal}
Junteng Liu, Shiqi Chen, Yu~Cheng, and Junxian He.
\newblock On the universal truthfulness hyperplane inside llms.
\newblock \emph{arXiv preprint arXiv:2407.08582}, 2024{\natexlab{c}}.

\bibitem[Liu et~al.(2023)Liu, Casper, Hadfield-Menell, and Andreas]{liu2023cognitive}
Kevin Liu, Stephen Casper, Dylan Hadfield-Menell, and Jacob Andreas.
\newblock Cognitive dissonance: Why do language model outputs disagree with internal representations of truthfulness?
\newblock In \emph{Proceedings of the 2023 Conference on Empirical Methods in Natural Language Processing}, pp.\  4791--4797, 2023.

\bibitem[Liu et~al.(2024{\natexlab{d}})Liu, Lin, Hewitt, Paranjape, Bevilacqua, Petroni, and Liang]{liu2024lost}
Nelson~F Liu, Kevin Lin, John Hewitt, Ashwin Paranjape, Michele Bevilacqua, Fabio Petroni, and Percy Liang.
\newblock Lost in the middle: How language models use long contexts.
\newblock \emph{Transactions of the Association for Computational Linguistics}, 12:\penalty0 157--173, 2024{\natexlab{d}}.

\bibitem[Lyu et~al.(2024)Lyu, Shridhar, Malaviya, Zhang, Elazar, Tandon, Apidianaki, Sachan, and Callison-Burch]{lyu2024calibrating}
Qing Lyu, Kumar Shridhar, Chaitanya Malaviya, Li~Zhang, Yanai Elazar, Niket Tandon, Marianna Apidianaki, Mrinmaya Sachan, and Chris Callison-Burch.
\newblock Calibrating large language models with sample consistency.
\newblock \emph{arXiv preprint arXiv:2402.13904}, 2024.

\bibitem[Mahaut et~al.(2024)Mahaut, Aina, Czarnowska, Hardalov, M{\"u}ller, and Marquez]{mahaut2024factual}
Mat{\'e}o Mahaut, Laura Aina, Paula Czarnowska, Momchil Hardalov, Thomas M{\"u}ller, and Lluis Marquez.
\newblock Factual confidence of {LLM}s: on reliability and robustness of current estimators.
\newblock In Lun-Wei Ku, Andre Martins, and Vivek Srikumar (eds.), \emph{Proceedings of the 62nd Annual Meeting of the Association for Computational Linguistics (Volume 1: Long Papers)}, pp.\  4554--4570, 2024.

\bibitem[Malinin \& Gales(2021)Malinin and Gales]{malinin2020uncertainty}
Andrey Malinin and Mark Gales.
\newblock Uncertainty estimation in autoregressive structured prediction.
\newblock In \emph{The Ninth International Conference on Learning Representations}, 2021.

\bibitem[Manakul et~al.(2023)Manakul, Liusie, and Gales]{manakul2023selfcheckgpt}
Potsawee Manakul, Adian Liusie, and Mark Gales.
\newblock Selfcheckgpt: Zero-resource black-box hallucination detection for generative large language models.
\newblock In \emph{Proceedings of the 2023 Conference on Empirical Methods in Natural Language Processing}, pp.\  9004--9017, 2023.

\bibitem[Marks \& Tegmark(2023)Marks and Tegmark]{marks2023geometry}
Samuel Marks and Max Tegmark.
\newblock The geometry of truth: Emergent linear structure in large language model representations of true/false datasets.
\newblock \emph{arXiv preprint arXiv:2310.06824}, 2023.

\bibitem[Mizrahi et~al.(2023)Mizrahi, Kaplan, Malkin, Dror, Shahaf, and Stanovsky]{mizrahi2023state}
Moran Mizrahi, Guy Kaplan, Dan Malkin, Rotem Dror, Dafna Shahaf, and Gabriel Stanovsky.
\newblock State of what art? a call for multi-prompt llm evaluation.
\newblock \emph{arXiv preprint arXiv:2401.00595}, 2023.

\bibitem[Naeini et~al.(2015)Naeini, Cooper, and Hauskrecht]{naeini2015obtaining}
Mahdi~Pakdaman Naeini, Gregory Cooper, and Milos Hauskrecht.
\newblock Obtaining well calibrated probabilities using bayesian binning.
\newblock In \emph{Proceedings of the AAAI conference on artificial intelligence}, volume~29, 2015.

\bibitem[Ni et~al.(2024)Ni, Bi, Guo, and Cheng]{ni2024llms}
Shiyu Ni, Keping Bi, Jiafeng Guo, and Xueqi Cheng.
\newblock When do {LLM}s need retrieval augmentation? mitigating {LLM}s{'} overconfidence helps retrieval augmentation.
\newblock In \emph{Findings of the Association for Computational Linguistics ACL 2024}, pp.\  11375--11388, 2024.

\bibitem[Nixon et~al.(2019)Nixon, Dusenberry, Zhang, Jerfel, and Tran]{nixon2019measuring}
Jeremy Nixon, Michael~W Dusenberry, Linchuan Zhang, Ghassen Jerfel, and Dustin Tran.
\newblock Measuring calibration in deep learning.
\newblock In \emph{CVPR workshops}, volume~2, 2019.

\bibitem[Ouyang et~al.(2022)Ouyang, Wu, Jiang, Almeida, Wainwright, Mishkin, Zhang, Agarwal, Slama, Ray, et~al.]{ouyang2022training}
Long Ouyang, Jeffrey Wu, Xu~Jiang, Diogo Almeida, Carroll Wainwright, Pamela Mishkin, Chong Zhang, Sandhini Agarwal, Katarina Slama, Alex Ray, et~al.
\newblock Training language models to follow instructions with human feedback.
\newblock \emph{Advances in neural information processing systems}, 35:\penalty0 27730--27744, 2022.

\bibitem[Patel et~al.(2021)Patel, Bhattamishra, and Goyal]{patel2021nlp}
Arkil Patel, Satwik Bhattamishra, and Navin Goyal.
\newblock Are nlp models really able to solve simple math word problems?
\newblock In \emph{Proceedings of the 2021 Conference of the North American Chapter of the Association for Computational Linguistics: Human Language Technologies}, pp.\  2080--2094, 2021.

\bibitem[Press et~al.(2023)Press, Zhang, Min, Schmidt, Smith, and Lewis]{press2023measuring}
Ofir Press, Muru Zhang, Sewon Min, Ludwig Schmidt, Noah~A Smith, and Mike Lewis.
\newblock Measuring and narrowing the compositionality gap in language models.
\newblock In \emph{Findings of the Association for Computational Linguistics: EMNLP 2023}, pp.\  5687--5711, 2023.

\bibitem[Rafailov et~al.(2024)Rafailov, Sharma, Mitchell, Manning, Ermon, and Finn]{rafailov2024direct}
Rafael Rafailov, Archit Sharma, Eric Mitchell, Christopher~D Manning, Stefano Ermon, and Chelsea Finn.
\newblock Direct preference optimization: Your language model is secretly a reward model.
\newblock \emph{Advances in Neural Information Processing Systems}, 36, 2024.

\bibitem[Rajpurkar et~al.(2016)Rajpurkar, Zhang, Lopyrev, and Liang]{rajpurkar2016squad}
Pranav Rajpurkar, Jian Zhang, Konstantin Lopyrev, and Percy Liang.
\newblock Squad: 100,000+ questions for machine comprehension of text.
\newblock In \emph{Proceedings of the 2016 Conference on Empirical Methods in Natural Language Processing}, pp.\  2383--2392, 2016.

\bibitem[Ram{\'\i}rez et~al.(2024)Ram{\'\i}rez, Birch, and Titov]{ramirez2024optimising}
Guillem Ram{\'\i}rez, Alexandra Birch, and Ivan Titov.
\newblock Optimising calls to large language models with uncertainty-based two-tier selection.
\newblock \emph{arXiv preprint arXiv:2405.02134}, 2024.

\bibitem[Reimers \& Gurevych(2019)Reimers and Gurevych]{reimers2019sentence}
Nils Reimers and Iryna Gurevych.
\newblock Sentence-bert: Sentence embeddings using siamese bert-networks.
\newblock In \emph{Proceedings of the 2019 Conference on Empirical Methods in Natural Language Processing and the 9th International Joint Conference on Natural Language Processing (EMNLP-IJCNLP)}, pp.\  3982--3992, 2019.

\bibitem[Ren et~al.(2023{\natexlab{a}})Ren, Luo, Zhao, Krishna, Saleh, Lakshminarayanan, and Liu]{ren2023outofdistribution}
Jie Ren, Jiaming Luo, Yao Zhao, Kundan Krishna, Mohammad Saleh, Balaji Lakshminarayanan, and Peter~J Liu.
\newblock Out-of-distribution detection and selective generation for conditional language models.
\newblock In \emph{The Eleventh International Conference on Learning Representations}, 2023{\natexlab{a}}.

\bibitem[Ren et~al.(2023{\natexlab{b}})Ren, Zhao, Vu, Liu, and Lakshminarayanan]{ren2023self}
Jie Ren, Yao Zhao, Tu~Vu, Peter~J Liu, and Balaji Lakshminarayanan.
\newblock Self-evaluation improves selective generation in large language models.
\newblock In \emph{Proceedings on "I Can't Believe It's Not Better: Failure Modes in the Age of Foundation Models" at NeurIPS 2023 Workshops}, pp.\  49--64, 2023{\natexlab{b}}.

\bibitem[Schulman et~al.(2017)Schulman, Wolski, Dhariwal, Radford, and Klimov]{schulman2017proximal}
John Schulman, Filip Wolski, Prafulla Dhariwal, Alec Radford, and Oleg Klimov.
\newblock Proximal policy optimization algorithms.
\newblock \emph{arXiv preprint arXiv:1707.06347}, 2017.

\bibitem[Sclar et~al.(2024)Sclar, Choi, Tsvetkov, and Suhr]{sclar2024quantifying}
Melanie Sclar, Yejin Choi, Yulia Tsvetkov, and Alane Suhr.
\newblock Quantifying language models' sensitivity to spurious features in prompt design or: How i learned to start worrying about prompt formatting.
\newblock In \emph{The Twelfth International Conference on Learning Representations}, 2024.

\bibitem[Sharma et~al.(2024)Sharma, Tong, Korbak, Duvenaud, Askell, Bowman, DURMUS, Hatfield-Dodds, Johnston, Kravec, Maxwell, McCandlish, Ndousse, Rausch, Schiefer, Yan, Zhang, and Perez]{sharma2024towards}
Mrinank Sharma, Meg Tong, Tomasz Korbak, David Duvenaud, Amanda Askell, Samuel~R. Bowman, Esin DURMUS, Zac Hatfield-Dodds, Scott~R Johnston, Shauna~M Kravec, Timothy Maxwell, Sam McCandlish, Kamal Ndousse, Oliver Rausch, Nicholas Schiefer, Da~Yan, Miranda Zhang, and Ethan Perez.
\newblock Towards understanding sycophancy in language models.
\newblock In \emph{The Twelfth International Conference on Learning Representations}, 2024.

\bibitem[Shi et~al.(2024)Shi, Han, Lewis, Tsvetkov, Zettlemoyer, and Yih]{shi2024trusting}
Weijia Shi, Xiaochuang Han, Mike Lewis, Yulia Tsvetkov, Luke Zettlemoyer, and Wen-tau Yih.
\newblock Trusting your evidence: Hallucinate less with context-aware decoding.
\newblock In \emph{Proceedings of the 2024 Conference of the North American Chapter of the Association for Computational Linguistics: Human Language Technologies (Volume 2: Short Papers)}, pp.\  783--791, 2024.

\bibitem[Si et~al.(2023{\natexlab{a}})Si, Gan, Yang, Wang, Wang, Boyd-Graber, and Wang]{si2022prompting}
Chenglei Si, Zhe Gan, Zhengyuan Yang, Shuohang Wang, Jianfeng Wang, Jordan Boyd-Graber, and Lijuan Wang.
\newblock Prompting gpt-3 to be reliable.
\newblock In \emph{The Eleventh International Conference on Learning Representations}, 2023{\natexlab{a}}.

\bibitem[Si et~al.(2023{\natexlab{b}})Si, Shi, Zhao, Zettlemoyer, and Boyd-Graber]{si2023getting}
Chenglei Si, Weijia Shi, Chen Zhao, Luke Zettlemoyer, and Jordan Boyd-Graber.
\newblock Getting more out of mixture of language model reasoning experts.
\newblock In \emph{Findings of the Association for Computational Linguistics: EMNLP 2023}, pp.\  8234--8249, 2023{\natexlab{b}}.

\bibitem[Stengel-Eskin et~al.(2024)Stengel-Eskin, Hase, and Bansal]{stengel2024lacie}
Elias Stengel-Eskin, Peter Hase, and Mohit Bansal.
\newblock Lacie: Listener-aware finetuning for confidence calibration in large language models.
\newblock \emph{arXiv preprint arXiv:2405.21028}, 2024.

\bibitem[Sun et~al.(2024)Sun, Huang, Wang, Wu, Zhang, Gao, Huang, Lyu, Zhang, Li, et~al.]{sun2024trustllm}
Lichao Sun, Yue Huang, Haoran Wang, Siyuan Wu, Qihui Zhang, Chujie Gao, Yixin Huang, Wenhan Lyu, Yixuan Zhang, Xiner Li, et~al.
\newblock Trustllm: Trustworthiness in large language models.
\newblock \emph{arXiv preprint arXiv:2401.05561}, 2024.

\bibitem[Team et~al.(2023)Team, Anil, Borgeaud, Wu, Alayrac, Yu, Soricut, Schalkwyk, Dai, Hauth, et~al.]{team2023gemini}
Gemini Team, Rohan Anil, Sebastian Borgeaud, Yonghui Wu, Jean-Baptiste Alayrac, Jiahui Yu, Radu Soricut, Johan Schalkwyk, Andrew~M Dai, Anja Hauth, et~al.
\newblock Gemini: a family of highly capable multimodal models.
\newblock \emph{arXiv preprint arXiv:2312.11805}, 2023.

\bibitem[Thirukovalluru et~al.(2024)Thirukovalluru, Huang, and Dhingra]{thirukovalluru2024atomic}
Raghuveer Thirukovalluru, Yukun Huang, and Bhuwan Dhingra.
\newblock Atomic self-consistency for better long form generations.
\newblock \emph{arXiv preprint arXiv:2405.13131}, 2024.

\bibitem[Thirunavukarasu et~al.(2023)Thirunavukarasu, Ting, Elangovan, Gutierrez, Tan, and Ting]{thirunavukarasu2023large}
Arun~James Thirunavukarasu, Darren Shu~Jeng Ting, Kabilan Elangovan, Laura Gutierrez, Ting~Fang Tan, and Daniel Shu~Wei Ting.
\newblock Large language models in medicine.
\newblock \emph{Nature medicine}, 29\penalty0 (8):\penalty0 1930--1940, 2023.

\bibitem[Tian et~al.(2023)Tian, Mitchell, Zhou, Sharma, Rafailov, Yao, Finn, and Manning]{tian2023just}
Katherine Tian, Eric Mitchell, Allan Zhou, Archit Sharma, Rafael Rafailov, Huaxiu Yao, Chelsea Finn, and Christopher~D Manning.
\newblock Just ask for calibration: Strategies for eliciting calibrated confidence scores from language models fine-tuned with human feedback.
\newblock In \emph{Proceedings of the 2023 Conference on Empirical Methods in Natural Language Processing}, pp.\  5433--5442, 2023.

\bibitem[Tian et~al.(2024)Tian, Mitchell, Yao, Manning, and Finn]{tian2024finetuning}
Katherine Tian, Eric Mitchell, Huaxiu Yao, Christopher~D Manning, and Chelsea Finn.
\newblock Fine-tuning language models for factuality.
\newblock In \emph{The Twelfth International Conference on Learning Representations}, 2024.

\bibitem[Tomani et~al.(2024)Tomani, Chaudhuri, Evtimov, Cremers, and Ibrahim]{tomani2024uncertainty}
Christian Tomani, Kamalika Chaudhuri, Ivan Evtimov, Daniel Cremers, and Mark Ibrahim.
\newblock Uncertainty-based abstention in llms improves safety and reduces hallucinations.
\newblock \emph{arXiv preprint arXiv:2404.10960}, 2024.

\bibitem[Touvron et~al.(2023)Touvron, Martin, Stone, Albert, Almahairi, Babaei, Bashlykov, Batra, Bhargava, Bhosale, et~al.]{touvron2023llama}
Hugo Touvron, Louis Martin, Kevin Stone, Peter Albert, Amjad Almahairi, Yasmine Babaei, Nikolay Bashlykov, Soumya Batra, Prajjwal Bhargava, Shruti Bhosale, et~al.
\newblock Llama 2: Open foundation and fine-tuned chat models.
\newblock \emph{arXiv preprint arXiv:2307.09288}, 2023.

\bibitem[Ulmer et~al.(2024)Ulmer, Gubri, Lee, Yun, and Oh]{ulmer2024calibrating}
Dennis Ulmer, Martin Gubri, Hwaran Lee, Sangdoo Yun, and Seong~Joon Oh.
\newblock Calibrating large language models using their generations only.
\newblock \emph{arXiv preprint arXiv:2403.05973}, 2024.

\bibitem[Varshney et~al.(2023)Varshney, Yao, Zhang, Chen, and Yu]{varshney2023stitch}
Neeraj Varshney, Wenlin Yao, Hongming Zhang, Jianshu Chen, and Dong Yu.
\newblock A stitch in time saves nine: Detecting and mitigating hallucinations of llms by validating low-confidence generation.
\newblock \emph{arXiv preprint arXiv:2307.03987}, 2023.

\bibitem[Wan et~al.(2024{\natexlab{a}})Wan, Huang, Cui, Quan, Bi, and Shi]{wan2024knowledge}
Fanqi Wan, Xinting Huang, Leyang Cui, Xiaojun Quan, Wei Bi, and Shuming Shi.
\newblock Knowledge verification to nip hallucination in the bud.
\newblock \emph{arXiv preprint arXiv:2401.10768}, 2024{\natexlab{a}}.

\bibitem[Wan et~al.(2024{\natexlab{b}})Wan, Wang, Liu, Alam, Zheng, Liu, Qu, Yan, Zhu, Zhang, Chowdhury, and Zhang]{wan2024efficient}
Zhongwei Wan, Xin Wang, Che Liu, Samiul Alam, Yu~Zheng, Jiachen Liu, Zhongnan Qu, Shen Yan, Yi~Zhu, Quanlu Zhang, Mosharaf Chowdhury, and Mi~Zhang.
\newblock Efficient large language models: A survey.
\newblock \emph{Transactions on Machine Learning Research}, 2024{\natexlab{b}}.
\newblock ISSN 2835-8856.

\bibitem[Wang et~al.(2023{\natexlab{a}})Wang, Yue, and Sun]{wang2023can}
Boshi Wang, Xiang Yue, and Huan Sun.
\newblock Can chatgpt defend its belief in truth? evaluating llm reasoning via debate.
\newblock In \emph{Findings of the Association for Computational Linguistics: EMNLP 2023}, pp.\  11865--11881, 2023{\natexlab{a}}.

\bibitem[Wang et~al.(2023{\natexlab{b}})Wang, Hu, Hou, Chen, Zheng, Wang, Yang, Huang, Ye, Geng, et~al.]{wang2023robustness}
Jindong Wang, Xixu Hu, Wenxin Hou, Hao Chen, Runkai Zheng, Yidong Wang, Linyi Yang, Haojun Huang, Wei Ye, Xiubo Geng, et~al.
\newblock On the robustness of chatgpt: An adversarial and out-of-distribution perspective.
\newblock \emph{arXiv preprint arXiv:2302.12095}, 2023{\natexlab{b}}.

\bibitem[Wang et~al.(2023{\natexlab{c}})Wang, Xu, Lan, Hu, Lan, Lee, and Lim]{wang2023plan}
Lei Wang, Wanyu Xu, Yihuai Lan, Zhiqiang Hu, Yunshi Lan, Roy Ka-Wei Lee, and Ee-Peng Lim.
\newblock Plan-and-solve prompting: Improving zero-shot chain-of-thought reasoning by large language models.
\newblock In \emph{Proceedings of the 61st Annual Meeting of the Association for Computational Linguistics (Volume 1: Long Papers)}, pp.\  2609--2634, 2023{\natexlab{c}}.

\bibitem[Wang et~al.(2024)Wang, Li, Feng, Yuan, Pan, Wang, Hu, and Li]{wang2024integrate}
Xinglin Wang, Yiwei Li, Shaoxiong Feng, Peiwen Yuan, Boyuan Pan, Heda Wang, Yao Hu, and Kan Li.
\newblock Integrate the essence and eliminate the dross: Fine-grained self-consistency for free-form language generation.
\newblock In Lun-Wei Ku, Andre Martins, and Vivek Srikumar (eds.), \emph{Proceedings of the 62nd Annual Meeting of the Association for Computational Linguistics (Volume 1: Long Papers)}, pp.\  11782--11794, 2024.

\bibitem[Wang et~al.(2023{\natexlab{d}})Wang, Wei, Schuurmans, Le, Chi, Narang, Chowdhery, and Zhou]{wang2023selfconsistency}
Xuezhi Wang, Jason Wei, Dale Schuurmans, Quoc~V Le, Ed~H. Chi, Sharan Narang, Aakanksha Chowdhery, and Denny Zhou.
\newblock Self-consistency improves chain of thought reasoning in language models.
\newblock In \emph{The Eleventh International Conference on Learning Representations}, 2023{\natexlab{d}}.

\bibitem[Wang et~al.(2023{\natexlab{e}})Wang, Li, Sun, and Liu]{wang2023self}
Yile Wang, Peng Li, Maosong Sun, and Yang Liu.
\newblock Self-knowledge guided retrieval augmentation for large language models.
\newblock In \emph{Findings of the Association for Computational Linguistics: EMNLP 2023}, pp.\  10303--10315, 2023{\natexlab{e}}.

\bibitem[Wei et~al.(2022)Wei, Wang, Schuurmans, Bosma, Xia, Chi, Le, Zhou, et~al.]{wei2022chain}
Jason Wei, Xuezhi Wang, Dale Schuurmans, Maarten Bosma, Fei Xia, Ed~Chi, Quoc~V Le, Denny Zhou, et~al.
\newblock Chain-of-thought prompting elicits reasoning in large language models.
\newblock \emph{Advances in neural information processing systems}, 35:\penalty0 24824--24837, 2022.

\bibitem[Wei et~al.(2023)Wei, Huang, Lu, Zhou, and Le]{wei2023simple}
Jerry Wei, Da~Huang, Yifeng Lu, Denny Zhou, and Quoc~V Le.
\newblock Simple synthetic data reduces sycophancy in large language models.
\newblock \emph{arXiv preprint arXiv:2308.03958}, 2023.

\bibitem[Wen et~al.(2024)Wen, Yao, Feng, Xu, Tsvetkov, Howe, and Wang]{wen2024art}
Bingbing Wen, Jihan Yao, Shangbin Feng, Chenjun Xu, Yulia Tsvetkov, Bill Howe, and Lucy~Lu Wang.
\newblock The art of refusal: A survey of abstention in large language models.
\newblock \emph{arXiv preprint arXiv:2407.18418}, 2024.

\bibitem[Williams et~al.(2018)Williams, Nangia, and Bowman]{williams2018broad}
Adina Williams, Nikita Nangia, and Samuel Bowman.
\newblock A broad-coverage challenge corpus for sentence understanding through inference.
\newblock In \emph{Proceedings of the 2018 Conference of the North American Chapter of the Association for Computational Linguistics: Human Language Technologies, Volume 1 (Long Papers)}, pp.\  1112--1122, 2018.

\bibitem[Wu et~al.(2016)Wu, Schuster, Chen, Le, Norouzi, Macherey, Krikun, Cao, Gao, Macherey, et~al.]{wu2016google}
Yonghui Wu, Mike Schuster, Zhifeng Chen, Quoc~V Le, Mohammad Norouzi, Wolfgang Macherey, Maxim Krikun, Yuan Cao, Qin Gao, Klaus Macherey, et~al.
\newblock Google's neural machine translation system: Bridging the gap between human and machine translation.
\newblock \emph{arXiv preprint arXiv:1609.08144}, 2016.

\bibitem[Xiao \& Wang(2021)Xiao and Wang]{xiao2021hallucination}
Yijun Xiao and William~Yang Wang.
\newblock On hallucination and predictive uncertainty in conditional language generation.
\newblock \emph{arXiv preprint arXiv:2103.15025}, 2021.

\bibitem[Xiao et~al.(2022)Xiao, Liang, Bhatt, Neiswanger, Salakhutdinov, and Morency]{xiao2022uncertainty}
Yuxin Xiao, Paul~Pu Liang, Umang Bhatt, Willie Neiswanger, Ruslan Salakhutdinov, and Louis-Philippe Morency.
\newblock Uncertainty quantification with pre-trained language models: A large-scale empirical analysis.
\newblock In \emph{Findings of the Association for Computational Linguistics: EMNLP 2022}, pp.\  7273--7284, 2022.

\bibitem[Xiong et~al.(2024)Xiong, Hu, Lu, LI, Fu, He, and Hooi]{xiong2024can}
Miao Xiong, Zhiyuan Hu, Xinyang Lu, YIFEI LI, Jie Fu, Junxian He, and Bryan Hooi.
\newblock Can llms express their uncertainty? an empirical evaluation of confidence elicitation in llms.
\newblock In \emph{The Twelfth International Conference on Learning Representations}, 2024.

\bibitem[Xu et~al.(2024{\natexlab{a}})Xu, Zhu, Ma, Zhang, Fan, Chen, and Yu]{xu2024rejection}
Hongshen Xu, Zichen Zhu, Da~Ma, Situo Zhang, Shuai Fan, Lu~Chen, and Kai Yu.
\newblock Rejection improves reliability: Training llms to refuse unknown questions using rl from knowledge feedback.
\newblock \emph{arXiv preprint arXiv:2403.18349}, 2024{\natexlab{a}}.

\bibitem[Xu et~al.(2024{\natexlab{b}})Xu, Wu, Diao, Liu, Wang, Chen, and Gao]{xu2024sayself}
Tianyang Xu, Shujin Wu, Shizhe Diao, Xiaoze Liu, Xingyao Wang, Yangyi Chen, and Jing Gao.
\newblock Sayself: Teaching llms to express confidence with self-reflective rationales.
\newblock \emph{arXiv preprint arXiv:2405.20974}, 2024{\natexlab{b}}.

\bibitem[Yadkori et~al.(2024)Yadkori, Kuzborskij, Stutz, Gy{\"o}rgy, Fisch, Doucet, Beloshapka, Weng, Yang, Szepesv{\'a}ri, et~al.]{yadkori2024mitigating}
Yasin~Abbasi Yadkori, Ilja Kuzborskij, David Stutz, Andr{\'a}s Gy{\"o}rgy, Adam Fisch, Arnaud Doucet, Iuliya Beloshapka, Wei-Hung Weng, Yao-Yuan Yang, Csaba Szepesv{\'a}ri, et~al.
\newblock Mitigating llm hallucinations via conformal abstention.
\newblock \emph{arXiv preprint arXiv:2405.01563}, 2024.

\bibitem[Yang et~al.(2024)Yang, Chen, and Pitas]{yang2024just}
Adam Yang, Chen Chen, and Konstantinos Pitas.
\newblock Just rephrase it! uncertainty estimation in closed-source language models via multiple rephrased queries.
\newblock \emph{arXiv preprint arXiv:2405.13907}, 2024.

\bibitem[Yang et~al.(2023)Yang, Chern, Qiu, Neubig, and Liu]{yang2023alignment}
Yuqing Yang, Ethan Chern, Xipeng Qiu, Graham Neubig, and Pengfei Liu.
\newblock Alignment for honesty.
\newblock \emph{arXiv preprint arXiv:2312.07000}, 2023.

\bibitem[Yang et~al.(2018)Yang, Qi, Zhang, Bengio, Cohen, Salakhutdinov, and Manning]{yang2018hotpotqa}
Zhilin Yang, Peng Qi, Saizheng Zhang, Yoshua Bengio, William Cohen, Ruslan Salakhutdinov, and Christopher~D Manning.
\newblock Hotpotqa: A dataset for diverse, explainable multi-hop question answering.
\newblock In \emph{Proceedings of the 2018 Conference on Empirical Methods in Natural Language Processing}, pp.\  2369--2380, 2018.

\bibitem[Yao et~al.(2024)Yao, Qi, Pan, Cao, Hu, Liu, Hou, and Li]{yao2024seakr}
Zijun Yao, Weijian Qi, Liangming Pan, Shulin Cao, Linmei Hu, Weichuan Liu, Lei Hou, and Juanzi Li.
\newblock Seakr: Self-aware knowledge retrieval for adaptive retrieval augmented generation.
\newblock \emph{arXiv preprint arXiv:2406.19215}, 2024.

\bibitem[Yin et~al.(2023)Yin, Sun, Guo, Wu, Qiu, and Huang]{yin2023large}
Zhangyue Yin, Qiushi Sun, Qipeng Guo, Jiawen Wu, Xipeng Qiu, and Xuan-Jing Huang.
\newblock Do large language models know what they don’t know?
\newblock In \emph{Findings of the Association for Computational Linguistics: ACL 2023}, pp.\  8653--8665, 2023.

\bibitem[Yue et~al.(2024)Yue, Zhao, Zhang, Du, and Yao]{yue2024large}
Murong Yue, Jie Zhao, Min Zhang, Liang Du, and Ziyu Yao.
\newblock Large language model cascades with mixture of thought representations for cost-efficient reasoning.
\newblock In \emph{The Twelfth International Conference on Learning Representations}, 2024.

\bibitem[Zhang et~al.(2024{\natexlab{a}})Zhang, Diao, Lin, Fung, Lian, Wang, Chen, Ji, and Zhang]{zhang2024r}
Hanning Zhang, Shizhe Diao, Yong Lin, Yi~Fung, Qing Lian, Xingyao Wang, Yangyi Chen, Heng Ji, and Tong Zhang.
\newblock R-tuning: Instructing large language models to say ‘i don’t know’.
\newblock In \emph{Proceedings of the 2024 Conference of the North American Chapter of the Association for Computational Linguistics: Human Language Technologies (Volume 1: Long Papers)}, pp.\  7106--7132, 2024{\natexlab{a}}.

\bibitem[Zhang et~al.(2024{\natexlab{b}})Zhang, Press, Merrill, Liu, and Smith]{zhang2024how}
Muru Zhang, Ofir Press, William Merrill, Alisa Liu, and Noah~A. Smith.
\newblock How language model hallucinations can snowball.
\newblock In \emph{Forty-first International Conference on Machine Learning}, 2024{\natexlab{b}}.

\bibitem[Zhang et~al.(2020)Zhang, Kishore, Wu, Weinberger, and Artzi]{bert-score}
Tianyi Zhang, Varsha Kishore, Felix Wu, Kilian~Q. Weinberger, and Yoav Artzi.
\newblock Bertscore: Evaluating text generation with bert.
\newblock In \emph{International Conference on Learning Representations}, 2020.

\bibitem[Zhang et~al.(2024{\natexlab{c}})Zhang, Peng, Tian, Zhou, Jin, Song, Mi, and Meng]{zhang2024self}
Xiaoying Zhang, Baolin Peng, Ye~Tian, Jingyan Zhou, Lifeng Jin, Linfeng Song, Haitao Mi, and Helen Meng.
\newblock Self-alignment for factuality: Mitigating hallucinations in {LLM}s via self-evaluation.
\newblock In \emph{Proceedings of the 62nd Annual Meeting of the Association for Computational Linguistics (Volume 1: Long Papers)}, pp.\  1946--1965. Association for Computational Linguistics, 2024{\natexlab{c}}.

\bibitem[Zhang et~al.(2023{\natexlab{a}})Zhang, Cui, Bi, and Shi]{zhang2023alleviating}
Yue Zhang, Leyang Cui, Wei Bi, and Shuming Shi.
\newblock Alleviating hallucinations of large language models through induced hallucinations.
\newblock \emph{arXiv preprint arXiv:2312.15710}, 2023{\natexlab{a}}.

\bibitem[Zhang et~al.(2023{\natexlab{b}})Zhang, Li, Cui, Cai, Liu, Fu, Huang, Zhao, Zhang, Chen, et~al.]{zhang2023siren}
Yue Zhang, Yafu Li, Leyang Cui, Deng Cai, Lemao Liu, Tingchen Fu, Xinting Huang, Enbo Zhao, Yu~Zhang, Yulong Chen, et~al.
\newblock Siren's song in the ai ocean: a survey on hallucination in large language models.
\newblock \emph{arXiv preprint arXiv:2309.01219}, 2023{\natexlab{b}}.

\bibitem[Zhang et~al.(2024{\natexlab{d}})Zhang, Li, Liu, Yu, Fung, Li, Li, and Ji]{zhang2024knowledge}
Yuji Zhang, Sha Li, Jiateng Liu, Pengfei Yu, Yi~R Fung, Jing Li, Manling Li, and Heng Ji.
\newblock Knowledge overshadowing causes amalgamated hallucination in large language models.
\newblock \emph{arXiv preprint arXiv:2407.08039}, 2024{\natexlab{d}}.

\bibitem[Zhao et~al.(2023)Zhao, Li, Joty, Qin, and Bing]{zhao2023verify}
Ruochen Zhao, Xingxuan Li, Shafiq Joty, Chengwei Qin, and Lidong Bing.
\newblock Verify-and-edit: A knowledge-enhanced chain-of-thought framework.
\newblock In \emph{Proceedings of the 61st Annual Meeting of the Association for Computational Linguistics (Volume 1: Long Papers)}, pp.\  5823--5840, 2023.

\bibitem[Zhao et~al.(2024)Zhao, Zhang, Pan, Yao, Yu, Wu, and Chen]{zhao2024fact}
Xinran Zhao, Hongming Zhang, Xiaoman Pan, Wenlin Yao, Dong Yu, Tongshuang Wu, and Jianshu Chen.
\newblock Fact-and-reflection ({F}a{R}) improves confidence calibration of large language models.
\newblock In Lun-Wei Ku, Andre Martins, and Vivek Srikumar (eds.), \emph{Findings of the Association for Computational Linguistics ACL 2024}, pp.\  8702--8718, 2024.

\bibitem[Zheng et~al.(2024)Zheng, Mishra, Chen, Cheng, Chi, Le, and Zhou]{zheng2024take}
Huaixiu~Steven Zheng, Swaroop Mishra, Xinyun Chen, Heng-Tze Cheng, Ed~H. Chi, Quoc~V Le, and Denny Zhou.
\newblock Take a step back: Evoking reasoning via abstraction in large language models.
\newblock In \emph{The Twelfth International Conference on Learning Representations}, 2024.

\bibitem[Zhou et~al.(2022)Zhou, He, Ma, Berg-Kirkpatrick, and Neubig]{zhou2022prompt}
Chunting Zhou, Junxian He, Xuezhe Ma, Taylor Berg-Kirkpatrick, and Graham Neubig.
\newblock Prompt consistency for zero-shot task generalization.
\newblock In \emph{Findings of the Association for Computational Linguistics: EMNLP 2022}, pp.\  2613--2626, 2022.

\bibitem[Zhou et~al.(2023{\natexlab{a}})Zhou, Sch{\"a}rli, Hou, Wei, Scales, Wang, Schuurmans, Cui, Bousquet, Le, and Chi]{zhou2023leasttomost}
Denny Zhou, Nathanael Sch{\"a}rli, Le~Hou, Jason Wei, Nathan Scales, Xuezhi Wang, Dale Schuurmans, Claire Cui, Olivier Bousquet, Quoc~V Le, and Ed~H. Chi.
\newblock Least-to-most prompting enables complex reasoning in large language models.
\newblock In \emph{The Eleventh International Conference on Learning Representations}, 2023{\natexlab{a}}.

\bibitem[Zhou et~al.(2023{\natexlab{b}})Zhou, Jurafsky, and Hashimoto]{zhou2023navigating}
Kaitlyn Zhou, Dan Jurafsky, and Tatsunori~B Hashimoto.
\newblock Navigating the grey area: How expressions of uncertainty and overconfidence affect language models.
\newblock In \emph{Proceedings of the 2023 Conference on Empirical Methods in Natural Language Processing}, pp.\  5506--5524, 2023{\natexlab{b}}.

\bibitem[Zhu et~al.(2023)Zhu, Xu, Wang, Zhang, and Mao]{zhu2023calibration}
Chiwei Zhu, Benfeng Xu, Quan Wang, Yongdong Zhang, and Zhendong Mao.
\newblock On the calibration of large language models and alignment.
\newblock In Houda Bouamor, Juan Pino, and Kalika Bali (eds.), \emph{Findings of the Association for Computational Linguistics: EMNLP 2023}, 2023.

\bibitem[Zhu \& Li(2023)Zhu and Li]{zhu2023physics}
Zeyuan~Allen Zhu and Yuanzhi Li.
\newblock Physics of language models: Part 3.1, knowledge storage and extraction.
\newblock \emph{arXiv preprint arXiv:2309.14316}, 2023.

\end{thebibliography}
\bibliographystyle{tmlr}


\end{document}